\documentclass[letterpaper]{article} % DO NOT CHANGE THIS
\usepackage{aaai23}  % DO NOT CHANGE THIS
\usepackage{times}  % DO NOT CHANGE THIS
\usepackage{helvet}  % DO NOT CHANGE THIS
\usepackage{courier}  % DO NOT CHANGE THIS
\usepackage[hyphens]{url}  % DO NOT CHANGE THIS
\usepackage{graphicx} % DO NOT CHANGE THIS
\usepackage{natbib}  % DO NOT CHANGE THIS AND DO NOT ADD ANY OPTIONS TO IT
\usepackage{caption} % DO NOT CHANGE THIS AND DO NOT ADD ANY OPTIONS TO IT
\usepackage{algorithm}
\usepackage{algorithmic}
\usepackage{amsmath, amssymb, mathtools}
\usepackage{multirow}
\usepackage{newfloat}
\usepackage{listings}
\DeclareCaptionStyle{ruled}{labelfont=normalfont,labelsep=colon,strut=off} % DO NOT CHANGE THIS
\floatstyle{ruled}
\newfloat{listing}{tb}{lst}{}
\floatname{listing}{Listing}
\title{CFFT-GAN: Cross-domain Feature Fusion Transformer for Exemplar-based Image Translation}
\author{
    Tianxiang Ma\textsuperscript{\rm 1,2}\equalcontrib,
    Bingchuan Li\textsuperscript{\rm 3},
    Wei Liu\textsuperscript{\rm 3},
    Miao Hua\textsuperscript{\rm 3},
    Jing Dong\textsuperscript{\rm 2}\thanks{Corresponding author.},
    Tieniu Tan\textsuperscript{\rm 2,4},
}
\affiliations{
    \textsuperscript{\rm 1}School of Artificial Intelligence, University of Chinese Academy of Sciences\\
    \textsuperscript{\rm 2}CRIPAC \& NLPR, Institute of Automation, Chinese Academy of Sciences\\
    \textsuperscript{\rm 3}ByteDance Ltd, Beijing, China \ \ 
    \textsuperscript{\rm 4}Nanjing University \ \ 
    
    tianxiang.ma@cripac.ia.ac.cn, \{libingchuan, huamiao\}@bytedance.com, liujikun63@gmail.com, \{jdong, tnt\}@nlpr.ia.ac.cn
}

\usepackage{bibentry}
\begin{document}

\maketitle

\begin{abstract}
Exemplar-based image translation refers to the task of generating images with the desired style, while conditioning on certain input image. Most of the current methods learn the correspondence between two input domains and lack the mining of information within the domains. In this paper, we propose a more general learning approach by considering two domain features as a whole and learning both inter-domain correspondence and intra-domain potential information interactions. Specifically, we propose a Cross-domain Feature Fusion Transformer (CFFT) to learn inter- and intra-domain feature fusion. Based on CFFT, the proposed CFFT-GAN works well on exemplar-based image translation. Moreover, CFFT-GAN is able to decouple and fuse features from multiple domains by cascading CFFT modules. We conduct rich quantitative and qualitative experiments on several image translation tasks, and the results demonstrate the superiority of our approach compared to state-of-the-art methods. Ablation studies show the importance of our proposed CFFT. Application experimental results reflect the potential of our method.

% 我们提出一种有别于之前基于卷积网络结构的CNN和ViT结合的图像翻译模型CFFT-GAN。我们提出基于ViT的特征融合模块CFFT，它能够更好的学习潜在特征之间的对应关系以及特征内部潜在的有利信息为了更有效的图像翻译。我们也探索了ViT在光流估计中的应用，提出形变版本模型CFFT-GAN-warp用来实现无配对数据人脸动画。我们在多种图像翻译任务中进行了丰富的定量和定型实验，结果表明了我们的方法相比SOTA方法的优越性。消融实验展示出我们CFFT的价值以及适用性。高分辨率图像生成结果体现出我们方法的应用潜力。

\end{abstract}

\begin{figure*}[t]
\centering
\includegraphics[width=1.0\linewidth]{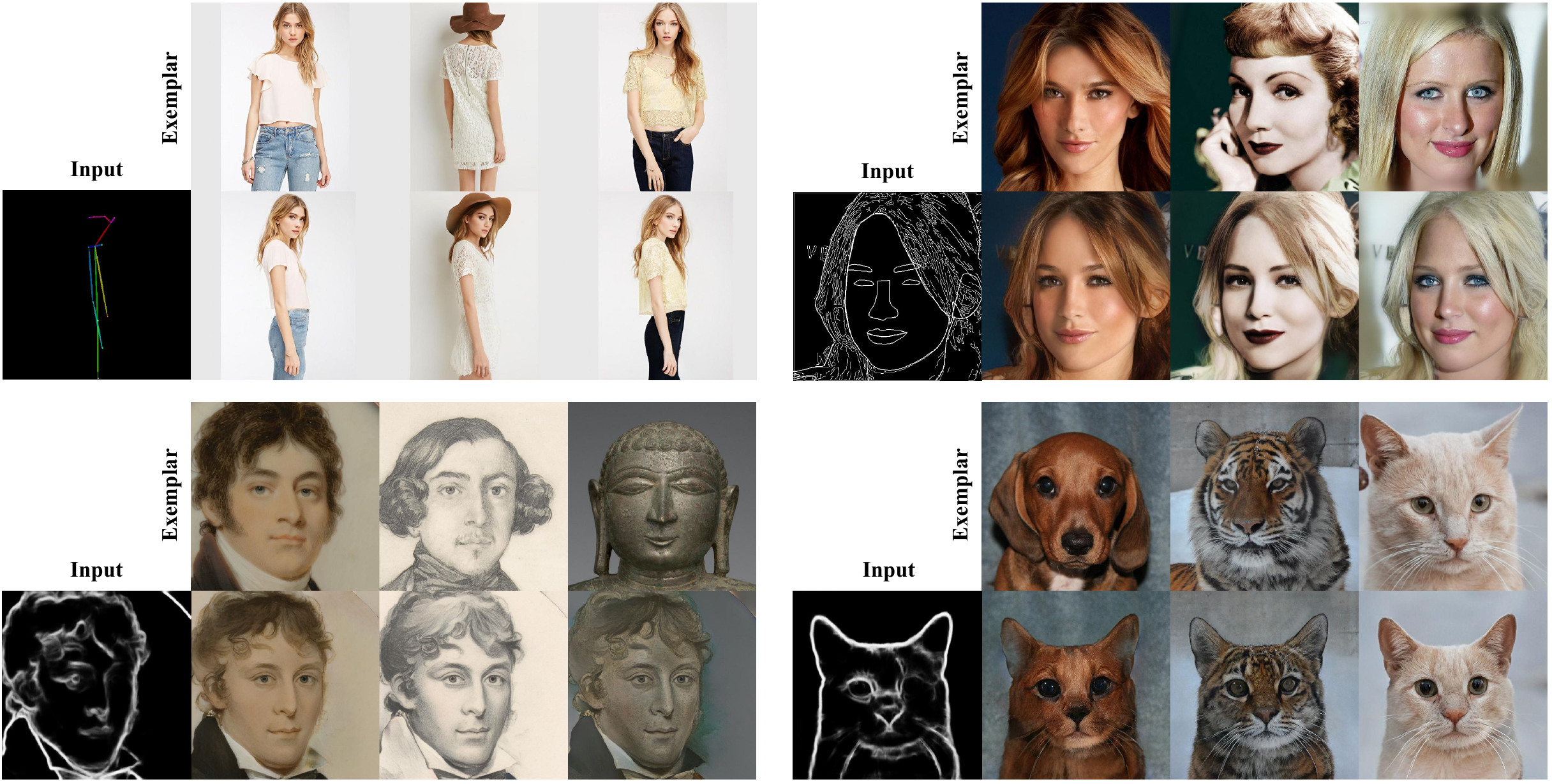}
\caption{Exemplar-based image translation results of our method on Deepfashion dataset, CelebA-HQ dataset, MetFace dataset, and AFHQ dataset.}
\label{fig:show}
\end{figure*}
 
\section{Introduction}
% Image-to-image translation aims to transform images between different domains and shows a wide range of applications \cite{choi2018stargan,isola2017image,park2020swapping,murez2018image}.

Exemplar-based image translation is a subdivision of image translation, which aims to generate a image with exemplar image style, while maintaining the content information of the input image. Such methods have a wide range of promising applications, such as style transfer, face editing, scene transformation, etc. 

% CoCosNet \cite{zhang2020cross} proposes a correspondence learning method by computing the correlation matrix of two domain features, which is essentially learning cross-attention. Other methods also learn the correspondence between two domains by various means.

%  use convolutional networks to learn the correspondence between domains, which has the limitation of the local receptive field. By contrast, the transformer is able to learn stronger global features through a self-attentive mechanism.

Previous extensive studies \cite{ma2018exemplar,wang2019example,huang2017arbitrary,park2019semantic,zhang2020cross,zhou2021cocosnet,zhan2021unbalanced} have made great progress in exemplar-based image translation. However, current methods mostly focus on learning the correspondence between the two input domains and ignore the potentially useful information interactions within each domain. In this paper, we propose a more general approach that combines the two domain features of the inputs into a whole, and then uses a transformer-based model to learn both the inter-domain correspondence and the information interactions within each domain. 

Recently, a number of transformer-based models \cite{dosovitskiy2020vit,carion2020end,wang2020axial,liu2021swin} have been applied to vision tasks. These works outperform CNN-based methods in their respective fields. Nevertheless, training a pure transformer model is time-consuming and highly dependent on the amount of data. Therefore, we utilize a hybrid structure of CNN and transformer to generate better results with limited amount of data. We first use CNN-based encoders to obtain features of the input image and the exemplar image and concatenate them into a whole. Then we propose a novel Cross-domain Feature Fusion Transformer (CFFT) to fuse the whole feature. The CFFT consists of a Feature Fusion Network and a Hierarchical Transformer Network. The former is used for feature fusion and the latter for learning fine-grained spatial alignment at multiple scales. By the improved transformer-based network, CFFT is able to continuously learn self-attention and cross-attention between two input features, so that the model can learn better inter-domain correspondence and potentially beneficial correlation information within each domain.

% The CFFT involves self-attention to learn intra-domain information and enables better interaction of information between the two domains.

% In the case of a limited amount of data, the hybrid structure of CNN and transformer is easier to generate better results \cite{esser2021taming}. 

% due to the limitation of the convolutional local receptive field and the computational complexity of the cross-attention map, it is difficult for them to fully utilize the global features of the label and example domain images, and at the same time, it is difficult to compute the cross-attention map between the two domains to the extreme.

% To give the model a deep understanding of the correspondence between content input and style exemplar domains, 

Based on the CFFT model, we propose CFFT-GAN, a general exemplar-based image translation method. It decouples and fuses features from different domains and leverages generative adversarial network to learn image translation. In particular, we employ encoders to extract features from different image domains and perform feature fusion through CFFT. Then, the fused feature is fed into the spatially-adaptive generator to synthesize the translation results. In addition, our CFFT model is a pluggable design, and we can achieve multi-domain image translation by cascading CFFT modules. 

We validate the different types of exemplar-based image translation capabilities of our method on many high-resolution image datasets, including pose-to-image, edge-to-image, sketch-to-image, and mask-to-image. These adequate and diverse experiments have demonstrated the validity of our model and show that our approach has better performance than state-of-the-art methods. Ablation studies and feature visualization results indicate the importance of our proposed CFFT module. Extended multi-domain experiments reflect the potential of our approach.

Our contributions can be summarized as follows:
\begin{itemize}
    % \item We propose a general learning approach by considering different domain features as a whole and learning both inter-domain correspondence and intra-domain potential information interactions.
    
    \item We present a general exemplar-based image translation method, CFFT-GAN, which using Cross-domain Feature Fusion Transformer module to learn both intra-domain and inter-domain correlation for more effective image translation.
    
    \item We propose a Hierarchical Transformer (Hiformer) network in CFFT for fine-grained spatial alignment at multiple scales.
   
    \item Benefiting from the pluggable nature of CFFT, we utilize cascaded CFFT modules to achieve multi-domain image translation.
   
    \item Rich validation and comparison experiments show that CFFT-GAN can achieve remarkable results in various image translation tasks and also demonstrate the superiority of our approach over SOTA methods.
    
    % \item We propose a Cross-domain Feature Fusion Transformer model, which consists of a Feature Fusion Network and a Spatial-aware Feature Self-learning network. 
     
    % learns better the correspondence between different domains as well as the potentially beneficial correlation within each domain.
    
    %  that utilizes transformer to learn both intra-domain and inter-domain correlation for more effective image translation and can decouple and fuse information from multiple domains by cascading CFFT modules.
    
    % a self-attentive feature fusion module with transformer (CFFT) for domain transfer and propose a multi-domain progressive fusion method, CFFT-Cascade that can efficiently learn associations between multiple feature domains. 
    
\end{itemize}

\begin{figure*}[t]
\centering
\includegraphics[width=0.96\linewidth]{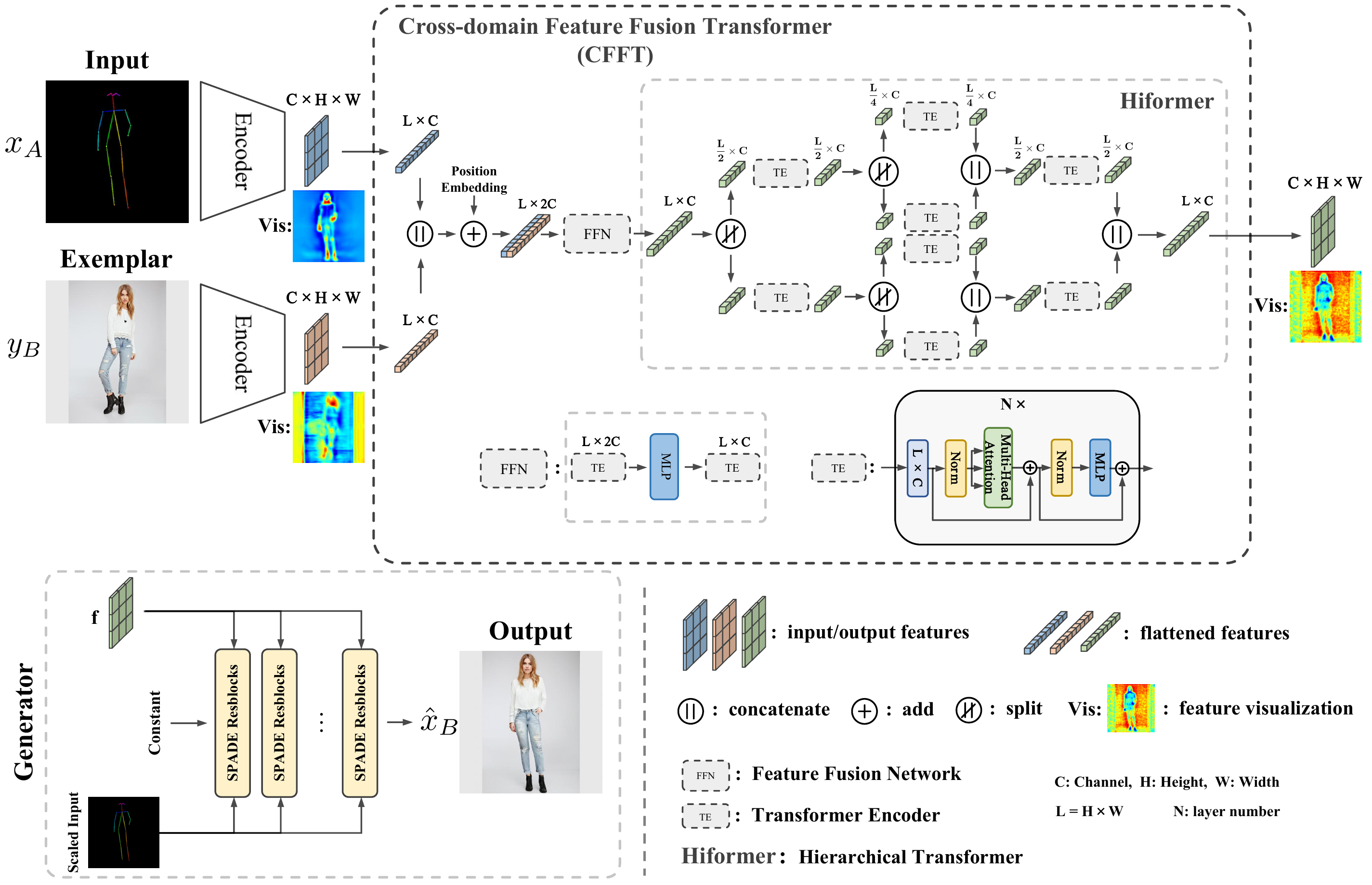}
\caption{Overall of our CFFT-GAN model for exemplar-based image translation. Two Encoders are used to extract different domain features. The CFFT module first concatenates different domain features as a whole feature, and performs feature fusion by FFN, and then learns fine-grained spatial alignment of the feature by Hiformer network. Finally, the output feature from CFFT is input to the spatially-adaptive generator to synthesize translation result.}

% The CFFT module learns the fusion of encoded features from both domains to obtain fused features $\mathbf{w}$. We then decode $\mathbf{w}$ by convolutional and upsampling layers to obtain multi-level fused features $\mathbf{w^{+}}$, and finally feed them and the conditional input to the SPADE residual network level by level to perform image translation.}
\label{fig:model}
\end{figure*}

\section{Related Work}

\subsection{Image Translation}
Image translation aims to learn the transformation between different image domains by training a conditional generative adversarial network. Some methods \cite{isola2017image,wang2018high,park2019semantic} use paired data for explicit supervision, while others \cite{zhu2017unpaired,choi2018stargan,choi2020stargan,yi2017dualgan,park2020contrastive} implement unsupervised training without paired data. Our method works well under both conditions. In addition image translation has been used in a variety of scenarios, such as image editing \cite{yu2018generative,wu2020cascade,shen2020interpreting}, domain adaptation \cite{liu2021source,li2020model}, pose transfer \cite{siarohin2018deformable,li2019dense,ren2020deep,ma2021must}. Recently, 
exemplar-based image translation \cite{ma2018exemplar,huang2018multimodal,wang2019example,saito2020coco,zhang2020cross,zhou2021cocosnet,zhan2021unbalanced} has received more and more attention because of its flexibility and applicability. Usually exemplar-based image translation refers to the transfer of stylistic features from a exemplar image to some type of label image, such as edges \cite{zhu2017toward,lee2018diverse}, key points \cite{ma2017pose,men2020controllable,ma2021must}, segmentation masks \cite{wang2018high,park2019semantic}, etc. Some approaches \cite{huang2018multimodal,men2020controllable,ma2021must} use AdaIN \cite{huang2017arbitrary} to transfer style information from the exemplar image into the label image. More recently, CoCosNet v1 \cite{zhang2020cross} uses correlation matrices to learn cross-domain correspondences. UNITE \cite{zhan2021unbalanced} introduces optimal transport for proper
feature alignment and faithful style control in image translation. CoCosNet v2 \cite{zhou2021cocosnet} learn a full-resolution correspondence with a hierarchical GRU-assisted PatchMatch \cite{barnes2009patchmatch}. In this paper, we propose a new method using the transformer-based network to learn cross-domain feature fusion. 

% Each layer of the our network can sense global information, similar to the correlation matrix computed in CoCosNet v1, but with better learnability.

% These methods use convolutional neural networks for learning latent features, whose local receptive fields limit the learning of long-range feature associations. Although the perceptual field can be increased to some extent by deepening the network, it is still not direct and effective enough.

\subsection{Transformer}
% In analogy to NLP tasks, the visual transformer model treats pixels or patches of an image as words, while the whole image looks at a sentence, which are then learned using the transformer model.

% then validated on various NLP tasks and quickly became a new generic network structure to replace traditional CNNs and RNNs.

Transformer was first proposed by \cite{vaswani2017attention} for machine translation. Early research on applying transformer to vision tasks has not received extensive attention, while the recent proposal of ViT \cite{dosovitskiy2020vit} model has set off a wave of research on visual transformer models. The idea of ViT is very straightforward, which is to transfer the original transformer model as invariant as possible to the image classification task. After ViT, more and more vision tasks started to use the transformer model and its variants. DETR \cite{carion2020end} applies transformer to object detection task with remarkable results. Axial-deeplab \cite{wang2020axial} applies axial attention to image segmentation. ViViT \cite{arnab2021vivit} proposes a video classification model based purely on visual transformer. Swin transformer \cite{liu2021swin} improves the performance of transformer for various visual tasks by further improving the computation of image patch window. Although transformer has achieved good results in various classical vision tasks, there are relatively few studies related to the use of transformer in dense estimation tasks like image generation. TransGAN \cite{jiang2021transgan} uses a pure transformer structure to design the generator and discriminator to train the adversarial model, but the resolution of the generated images is small. Recently, ViTGAN \cite{lee2021vitgan} combines ViT and GAN and designs a new regularization technique to achieve comparable results to CNN-based GAN models on many standard image generation benchmarks. These studies demonstrate that transformer models can also play an effective role in image generation. However, the mainstream image translation approaches are still using CNN-based networks. Because image translation requires that the generated image has accurate semantic information about the conditional input as well as style information about the exemplar image, which demands learning the correspondence between different image domains. In this paper, we propose a cross-domain feature fusion transformer model for exemplar-based image translation. This model can better learn the correspondences between different domains and the potential connections within each domain.

% In addition, VQGAN \cite{esser2021taming} does not use transformer structure completely, but uses a combination of CNN and transformer to achieve good generation results as well.

% transformer最早是由A提出的用于机器翻译的模型，之后在各种NLP任务上得到验证并迅速成为一种代替传统CNN和RNN的新型通用网络结构。视觉transformer模型是类比NLP任务将图像的像素或patch看作单词，而整个图像看着句子，然后使用transformer模型进行学习。早期将transformer迁移到视觉任务的研究没有得到广泛的关注，而最近ViT模型的提出掀起了视觉transformer模型的研究热潮。ViT的思想很直接，就是将原始的transformer模型尽可能不变的迁移到图像分类任务中，将一幅图像中每个16*16大小像素块看作是一个单词，并提出position embedding和class token的有效实现方式，最后像CNN一样堆叠transformer模块来更好的学习分类特征。

% 在ViT之后，越来越多的视觉任务开始使用transfomrer模型和它的变种。DETR将trannformer应用到目标检测任务中取得了显著效果。Axial-deeplab将axial attention应用到图像分割中。ViViT提出一种纯基于视觉transformer的视频分类模型。Swin transformer改进了图像patch窗口的计算方式进一步提升了transformer在各种视觉任务上的性能。

% 尽管transformer已经在各种经典视觉任务中取得了很好的结果，但在图像生成这种稠密估计任务中使用transformer的相关研究还比较少。TransGAN使用纯transformer结构设计生成器和判别器来训练对抗模型，但生成图像的分辨率不高。最近，ViTGAN将ViT模型和生成对抗网络结合，并设计新的正则化技术，在许多标准图像生成基准上取得和基于CNN的GAN模型相当的效果。VQGAN没有完全使用transformer结果，而是采用CNN和transformer相结合的方式也同样实现不错的生成效果。这些研究展示出视觉transformer模型在图像生成任务中也能发挥有效的作用。对于图像翻译任务来说目前主流方法还都是采用CNN结构的网络模型，因为图像翻译要求生成的目标图像具有关于输入label准确的的语义信息，这需要学习不同图像域之间的对应关系。在本文中，我们提出一种新颖的基于视觉transformer的图像翻译模型，这将进一步扩展transformer在图像生成领域的应用前景。

% \subsection{Optical Flow Estimation}

\section{Approach}
% （总起段）说明任务setting和一些变量字母设置，总的概括整体模型，简单说明接下来几段各种的主要内容。
Exemplar-based image translation refers to transfer an input image $x_{A}$ from domain $A$ to domain $B$, and make it have the stylistic texture of exemplar $y_{B}$, while maintaining the content of $x_{A}$. Our proposed CFFT-GAN first transform $x_{A}$ and $y_{B}$ into feature space $F$ by two CNN-based encoders, then fuse them using Cross-domain Feature Fusion Transformer, and generate the translation result via a SPADE-based \cite{park2019semantic} residual generator. The overall of our method is illustrated in Figure \ref{fig:model}.

% 基于模版的图像翻译旨在将域A中的图像xA迁移到域B中，并且使得其具有yB的风格纹理，同时保持xA的语义。为此，我们提出一种新颖的模型，CFFT-GAN。我们首先通过两个编码器分别将xA和yB转化到一个对齐特征域S中，然后利用我们提出的CFFT模块对它们进行特征融合，最后通过翻译网络从融合特征w中生成最终结果。图A中展示出模型的整体结构。此外，我们还提出了该模型的形变版本，CFFT-GAN-warp，它可以实现无配对训练数据条件下的人脸动画。

% \subsection{Learning and Alignment of Domain Features}
\subsection{Encoders}
\label{subsection1}
For exemplar-based image translation, the inputs are two different types of image domains, content input domain $A$ and style exemplar domain $B$. There is large semantic gaps between them, which is not conducive to the feature fusion later. To learn correspondence better, we need to map the two image domains into the same latent feature space $F$. Specifically, we let $\mathcal{E}_{A \rightarrow S}$ and $\mathcal{E}_{B \rightarrow S}$ represent the encoders of two domains, then the domain aligned latent features can be expressed as:
\begin{equation}
\begin{aligned}
&x_{F}=\mathcal{E}_{A \rightarrow F}\left(x_{A} ; \theta_{\mathcal{E_{A}}}\right), \\
&y_{F}=\mathcal{E}_{B \rightarrow F}\left(y_{B} ; \theta_{\mathcal{E_{B}}}\right),
\end{aligned}
\end{equation}
where $\theta$ is the learnable parameters for each encoder, and $x_{F}$ and $y_{S} \in \mathbb{R}^{C \times H \times W}$ represent the learned domain aligned features, which are used for later feature fusion. The domain alignment of features is trained by Equation \ref{domain_align}.

% In the process, we also constrain the label image to match its corresponding realistic image in the $F$ domain. Domain alignment facilitates the subsequent feature fusion learning.

% 对于基于模版的图像翻译来说，输入是两种不同类型的图像域，label域A和模版域B。它们之前可能存在着较大的语义差异，例如sketch图像和真实图像。为了能更好的学习他们之前的对应关系，我们希望将两种图像域映射到同一个语义特征空间S中来实现域对齐。同之前工作A一样，我们首先通过两个参数不同的可学习编码器网络来学习label图像域和模版图像域的潜在特征，并在这个过程中约束label图像域特征与其对应的真实图像的特征相一致，这就使得从两个不同域学到的潜在特征处于相同的语义特征空间中。这样有利于后续的特征融合学习。

% A limitation of transformer is that the number of model parameters can increase dramatically when the scale of the input features is too large, since it uses all MLP and multihead self-attention (MSA) to learn features. 

% Then we want to learn the feature fusion of $x_{S}$ and $y_{S}$. Previous studies have focused on learning the correspondence between them while neglecting the potentially useful information of each feature itself.
% the MSA provides self-attention for intra-domain features and MLP offer cross-attention for inter-domain features.

\subsection{Cross-domain Feature Fusion Transformer}
We propose a novel Cross-domain Feature Fusion Transformer (CFFT) model, shown on the Figure \ref{fig:model}. The CFFT is designed to learn not only inter-domain correspondences, but also feature correlations within domains. Therefore, we can more fully utilize the features of both domains for image translation. Cross-domain feature pairs are first flattened into feature vectors. Then we concatenate the flattened features into a whole along the channel dimension. There are a total of $L=H \times W$ tokens, and the dimension of each token is $1 \times 2C$. The purpose of this operation is that the transformer-based network can learn the correlation information within each feature and also indirectly learn the correspondence between different domain features.

\noindent \textbf{Feature Fusion Network.}
After adding the position embedding, we then use Feature Fusion Network (FFN) to learn feature fusion and semantic coarse matching. FFN consists of two Transformer Encoders (TE) \cite{vaswani2017attention}, each TE has a module layer number N=3. The two TEs are connected through a MLP layer for compressing feature dimension from 2C to C. In this way, FFN achieves a feature fusion role, and based on the transformer's self-attention mechanism, the module is able to learn both the inter-domain and intra-domain correlations within the total feature.

\noindent \textbf{Hierarchical Transformer.}
To decouple and match the two features more finely, we propose a Hierarchical Transformer (Hiformer) after FFN. It presents a stepped structure that continuously splits features in spatial dimension and performs self-learning optimization for each spatial part of the feature layer by layer. For each TE in Hiformer, we use a module layer number N=2. This approach implements a transformer-based feature learning process from coarse to fine. The number of feature splitting levels is adjustable, and we use 3 levels in this paper. We then gradually fuse the local features by same structure. 

With the above network structure, we can obtain the output feature that have been decoupled and spatially aligned, as shown in the feature visualizations of Figure \ref{fig:model}.

\begin{figure}[t]
\centering
\includegraphics[width=1.0\linewidth]{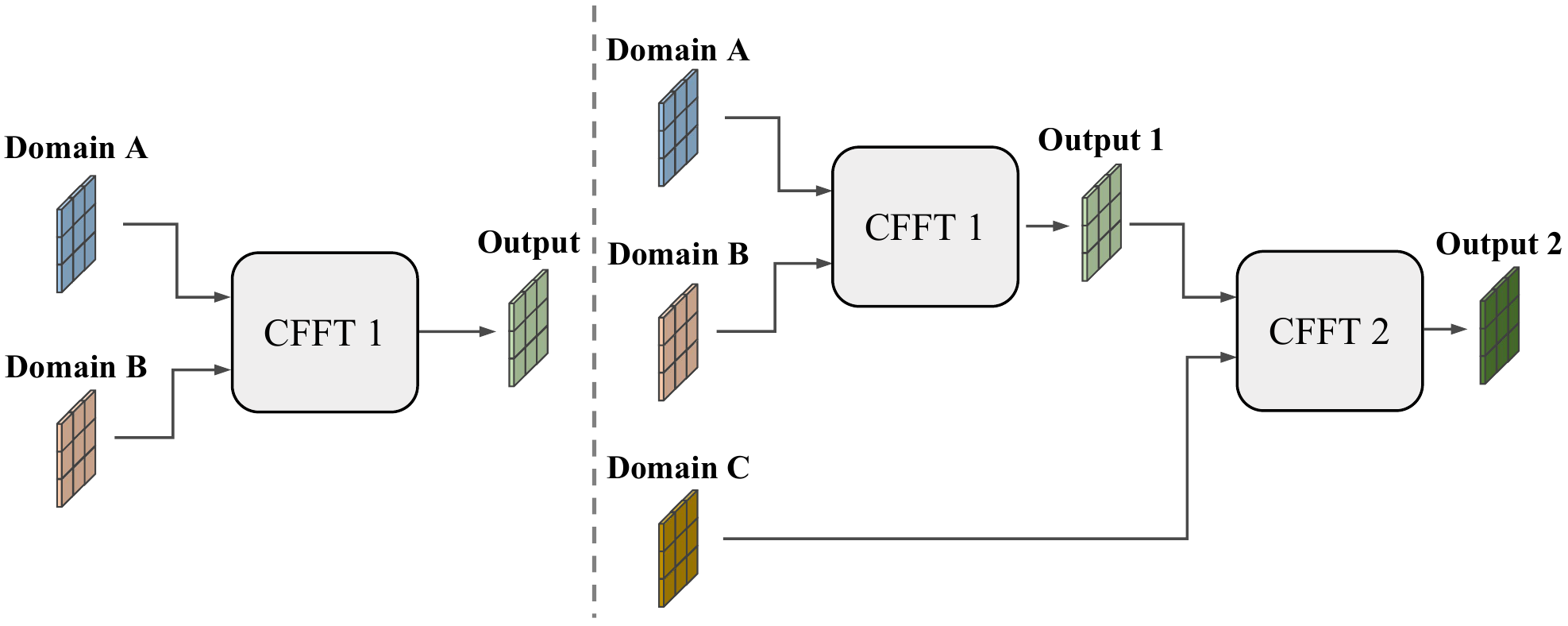}
\caption{Our method can achieve multi-domain image translation by cascading CFFT modules.}
\label{fig:model2}
\end{figure}

\subsection{Spatially-adaptive Generator}
The role of the generator network is to synthesize $\hat{x}_{B}$ with the content of input image and the style of exemplar image. In the previous step, we have obtained the fused and aligned feature named $\mathbf{f}$, we just need to feed it into the generator network. Therefore, we introduce a spatially-adaptive denormalization (SPADE) \cite{park2019semantic} structure as the generator. We concatenate the feature $\mathbf{f}$ with the scaled input image into $\mathbf{w}$, where scaled input image can help the initial training of the model. Then we feed $\mathbf{w}$ into the SPADE-based residual blocks as modulation parameters layer by layer. Specifically, for a particular feature layer $F^{i} \in \mathbb{R}^{C_{i} \times H_{i} \times W_{i}}$ in the generator, we have the following formula,
% first map $\mathbf{w}$ to the two modulation parameters $\alpha_{h, w}^{i}$ and $\beta_{h, w}^{i}$ by means of two convolutional layers, which are then used to implement the following denormalization operation:
\begin{equation}
F_{c, h, w}^{i} = \gamma_{h, w}^{i}(\mathbf{w}_{i}) \times \frac{F_{c, h, w}^{i}-\mu_{h, w}^{i}}{\sigma_{h, w}^{i}}+\beta_{h, w}^{i}(\mathbf{w}_{i}),
\end{equation}
where $\mu_{h, w}^{i}$ and $\sigma_{h, w}^{i}$ are the calculated mean and standard deviation across channel dimension, $\gamma_{h, w}^{i}$ and $\beta_{h, w}^{i}$ are learnable weight networks, $\mathbf{w}_{i}$ is a copy of $\mathbf{w}$. Finally, the generated result of the image translation can be expressed as:
\begin{equation}
\hat{x}_{B}=\mathcal{G}\left(\mathbf{w}; \theta_{\mathcal{G}}\right),
\end{equation}
where $\theta_{\mathcal{G}}$ is the learnable parameters of the generator $\mathcal{G}$.
% 翻译网络的作用是生成具有输入label图像语义结构且具有模版图像风格的xB。在上一步中我们已经得到了两个图像域特征融合后的新特征w，我们只需将它输入生成器网络中。与之前的工作类似，我们采用SPADE结构作为生成器主干，将w作为调制参数作用到各个层级的SPADE残差块中。具体的，对于生成器某一层特征A来说，我们首先通过两个卷积神经网络来将w映射为两个调制参数a和b，然后通过如下denormalization操作来迁移w特征到生成器主干特征上：

\subsection{Multi-domain Image Translation}
\label{CFFT-cascade}

Our proposed CFFT is a pluggable module design. When the inputs have more than two image domains, we can also easily decouple and fuse information from multiple domains by cascading CFFT modules. As shown in schematic Figure \ref{fig:model2}, we first learn a feature fusion using CFFT1 for domains A and B, then introduce a third domain C and jointly learn a feature fusion for all three domains by CFFT2. This way of learning facilitates the progressive fusion of features, allowing for better correspondence learning between several different domains. As shown in Figure \ref{fig:animation}, our method can introduce a third domain (pose) on top of the dual-domain image translation to achieve both image translation and face animation.

\subsection{Loss Functions}
We train the CFFT-GAN model end-to-end. Usually, semantically aligned data for different domains is accessible, for example $x_{A}$ and $x_{B}$, whereas triplets data $(x_{A}$,$y_{B};x_{B})$ may not be accessible. We therefore construct the pseudo-tuple $(x_{A},\widetilde{x}_{B};x_{B})$, where $\widetilde{x}_{B}=\mathcal{A}(x_{B})$, $\mathcal{A}$ denotes a collection of data augmentation operations, such as horizontal flip, geometric deformation, random crop, etc. The following loss functions are used for the training process.

\noindent \textbf{Domain Alignment Loss.} We need the features $x_{F}$ and $y_{F}$ learned by the encoders to be in the same feature domain, as explained in subsection "Encoders". Therefore, we let different domain features under the same semantic be aligned.
\begin{equation}
\mathcal{L}_{\text {align}}=\left\|\mathcal{E}_{A \rightarrow S}\left(x_{A}\right)-\mathcal{E}_{B \rightarrow S}\left(x_{B}\right)\right\|_{1}.
\label{domain_align}
\end{equation}

\noindent \textbf{Feature Matching Loss.} Following previous work \cite{johnson2016perceptual,zhang2020cross}, we used the pre-trained VGG-19 network to extract the features of the generated translation result and the corresponding Ground Truth separately, and let the features at each layer match.
\begin{equation}
\mathcal{L}_{\text {match}}=\sum_{l} \mu_{l}\left\|\phi_{l}\left(\hat{x}_{B}\right)-\phi_{l}\left(x_{B}\right)\right\|_{1},
\end{equation}
where $\phi_{l}$ is the output features of layer $l$ in the VGG-19 network, and $\mu_{l}$ is the corresponding weight for each layer.

\noindent \textbf{Translation Loss.} We want the model to output translation image that has semantic content of the input image and the stylized texture of the exemplar image. Therefore, We utilize perceptual loss to reduce semantic differences and contextual loss \cite{mechrez2018contextual} to constrain the stylistic consistency. Specifically, the perceptual loss is shown as follows,
\begin{equation}
\mathcal{L}_{\text {perc}}=\left\|\phi_{h}\left(\hat{x}_{B}\right)-\phi_{h}\left(x_{B}\right)\right\|_{2},
\end{equation}
where $\phi_{h}$ represents the output of a high-level layer in the pre-trained VGG-19 network, and this layer contains mainly high-level semantic information about the input image. The contextual loss (CX) is written as follows,
\begin{equation}
\mathcal{L}_{CX}=\sum_{l} \omega_{l}\left[-\log \left(C X\left(\phi_{l}\left(\hat{x}_{B}\right), \phi_{l}\left(y_{B}\right)\right)\right)\right],
\end{equation}
where CX stands for contextual similarity, $\phi_{l}$ is the $l$th layer of low-level features in the pre-trained VGG-19 network, which contain mainly the rich stylized texture of the image, and $\omega_{l}$ is the weight coefficients of the different layers.

\noindent \textbf{Adversarial Loss.} We also use a discriminator network to distinguish between the real samples and the generated samples, and train it together with our image translation model. The adversarial loss function is defined as,
\begin{equation}
\begin{aligned}
\mathcal{L}_{a d v}^{\mathcal{D}} &=-\mathbb{E}\left[h\left(\mathcal{D}\left(y_{B}\right)\right)\right]-\mathbb{E}\left[h\left(-\mathcal{D}\left(\mathcal{G}\left(x_{A}, y_{B}\right)\right)\right)\right], \\
\mathcal{L}_{a d v}^{\mathcal{G}} &=-\mathbb{E}\left[\mathcal{D}\left(\mathcal{G}\left(x_{A}, y_{B}\right)\right)\right],
\end{aligned}
\end{equation}
where $h(t)=min(0,-1+t)$ represents the hinge function to regularize the discriminator.

\noindent \textbf{Total Loss.} In conclusion, we optimize the following overall objective.
\begin{equation}
\begin{aligned}
\mathcal{L} &= \lambda_{\text{align}} \mathcal{L}_{\text {align }}+\lambda_{\text{match}} \mathcal{L}_{\text {match}}+\lambda_{\text{perc}} \mathcal{L}_{\text{perc}} \\
&+\lambda_{\text{CX}}\mathcal{L}_{\text {CX}}+\lambda_{\text{adv}}\left(\mathcal{L}_{\text {adv }}^{\mathcal{D}}+\mathcal{L}_{\text {adv }}^{\mathcal{G}}\right),
\end{aligned}
\end{equation}
where weights $\lambda$ are used to balance the individual objectives.

% 我们端到端的训练CFFT-GAN模型。通常，不同域的语义对齐数据是容易获取的，例如A和B，而三元组数据（A，B，C）可能无法获取，因此我们构建伪三元组数据（A），其中  而对于CFFT-GAN-warp我们采用在CFFT-GAN预训练模型的基础上进行联合finetune。训练过程采用如下的损失函数。
% 域对齐损失：我们需要编码器学习到的潜在特征xS和yS在相同的特征域中，为此我们让同一个语义下的不同域特征进行对齐：

% 特征匹配损失：沿用之前的工作，我们使用预训练的VGG-19网络来分别提取生成的翻译结果和对应的Ground Truth的特征，并让各层级的特征相匹配。

% 翻译损失：我们希望模型输出的翻译结果在语义内容上与输入标签一致并且具有输入模版图的风格纹理。因此我们分别采用感知损失来减小语义差异以及上下文损失来约束翻译结果与模版的风格一致性。具体的，感知损失如下：

% 对抗损失: 我们也使用一个判别器网络来区别真实样本和生成的fake样本，并将它和我们的图像翻译模型一起对抗训练，对抗损失函数如下。

% 总损失

\begin{table*}[t]
\centering
\begin{tabular}{l|ccc|cc|cc|cc}
\hline
% \noalign{\smallskip}
& \multicolumn{3}{c|}{Deepfashion} & \multicolumn{2}{|c|}{CelebA-HQ} & \multicolumn{2}{|c|}{Metfaces} & \multicolumn{2}{|c}{ADE20K}\\
\cline{2-10} 
& \text { FID $\downarrow$ } & \text { SWD $\downarrow$ } & \text { LPIPS $\downarrow$ }  & \text { FID $\downarrow$ } & \text { SWD $\downarrow$ } & \text { FID $\downarrow$ } & \text { SWD $\downarrow$ } & \text { FID $\downarrow$ } & \text { SWD $\downarrow$ } \\
% \noalign{\smallskip}
\hline
% \noalign{\smallskip}
{CoCosNet v1}  & {14.4} & {29.0} & {0.2386} & {19.4} & {22.3} & {25.6} & { 24.3} & {26.4} & {10.5} \\
{UNITE}        & {13.1} & {26.6} & {0.2371} & {18.3} & {21.6} & {24.1} & {23.5} & \textbf{25.1} & {10.1}\\
{CoCosNet v2}  & {12.2} & {24.6} & {0.2245} & {16.2} & {19.7} & {23.3} & {22.4} & {25.2} & \textbf{9.9} \\
{Ours}   & \textbf{8.8} & \textbf{19.0} & {\textbf{0.2070}} & \textbf{12.4} & \textbf{18.1} & \textbf{22.6} & \textbf{21.3} & {25.4} & {10.1}\\
\hline
\end{tabular}
\caption{Quantitative comparison results of image quality between our method and SOTA methods.}
\label{table:metrics1}
\end{table*}

% The first column shows the different types of label input, the second column presents the ground truth corresponding to the label, the remaining columns represent the generated results obtained by each method, and the last column shows the reference style template diagram.

% Input edge and exemplar, our method can perform image translation, and then input head pose by cascading CFFT, the model can conduct face animation based on the previous image translation.

\section{Experiments}
% Deepfashion，CelebAHQ，MetFaces，ADE20K
\textbf{Datasets.} To verify the generality of our approach, we perform experiments on several widely used public datasets.
\begin{itemize}
    \item Deepfashion \cite{liu2016deepfashion} contains 52,712 human images with various appearances and poses. Human image paired data is used in our experiments. We also utilize Openpose \cite{cao2017realtime} to extract the key points of the human image as input pose.
    \item CelebA-HQ \cite{liu2015deep} is comprised of 30,000 high resolution images of faces in a variety of styles. We connect its face landmarks to form the face edge, and use the Canny edge detection algorithm to get the background edge, and finally combine them as the input.
    \item MetFaces \cite{karras2020training} is a high resolution art-style portrait dataset, which contains 1,336 images of human faces. We use HED \cite{xie2015holistically} to extract the sketch of the face images.
    \item ADE20K \cite{zhou2017scene} contains 25,574 scene images with 150-class semantic segmentation annotation. We use this dataset to validate the image translation from mask to scene.
    \item AFHQ \cite{choi2020stargan} is a animal faces dataset consisting of 15,000 high-quality images. We also use HED to extract the sketch of the animal faces.
\end{itemize}
For better comparison with the SOTA methods and to demonstrate the good performance of our method, all datasets are experimented with 512 resolution images, except for ADE20K where 256 resolution images are used.

\noindent \textbf{Baselines.} We compare our method with SOTA exemplar-based image translation methods. 1) CoCosNet v1 \cite{zhang2020cross} summarized various types of image translation tasks into exemplar-based image translation for the first time. It proposes a cross-domain correspondence learning method, which deforms the exemplar image by calculating the correlation matrix of two domain features. 2) UNITE \cite{zhan2021unbalanced} introduces optimal transport for proper feature alignment and faithful style control in image translation. 3) CoCosNet v2 \cite{zhou2021cocosnet} learn a full-resolution correspondence, and purpose a hierarchical GRU-assisted PatchMatch method for efficient correspondence computation. Because we propose a general image translation method, there is no comparison with task-specific methods. Besides, we do not compare with the approaches that directly learn image translation and fail to use an exemplar.

\noindent \textbf{Evaluation Metrics.} We adopt the commonly used image generation evaluation metrics to assess the performance of the various methods. Fréchet Inception Score (FID) \cite{heusel2017gans} is the metric used to evaluate the quality and diversity of generated images. Sliced Wasserstein distance (SWD) \cite{karras2017progressive} is applied to measure the statistical distance of low-level patch distribution between synthesized images and real images. Learned Perceptual Image Patch Similarity (LPIPS) is a semantic similarity evaluation metric, and we use it to evaluate the decoupling and semantic alignment of the models on Deepfashion dataset, since only Deepfashion has paired data to compute this metric.

For exemplar-based image translation, the generated image should have the same semantic content as the input and have the consistent style and texture as the exemplar. We therefore calculate two additional metrics for evaluating semantic consistency and style similarity. We utilize the pre-trained VGG model on ImageNet and use its high-level feature maps, $relu3\_ 2$, $relu4\_ 2$ and $relu5\_ 2$, to represent high-level semantics. We then calculate the high-level semantics average cosine similarity between the generated and content input images as a semantic consistency metric. Similarly, the style similarity metric can be calculated using the low-level feature maps, $relu1\_ 2$, $relu2\_ 2$.

\begin{table}[t]
\centering
\setlength{\tabcolsep}{0.4mm}{
\begin{tabular}{l|c|c|c|c}
\hline
% \noalign{\smallskip}
& {Deepfashion} & {CelebA-HQ} & {Metfaces} & {ADE20K}\\

\hline
% \noalign{\smallskip}
{CoCosNet v1}  & {0.924} & {0.945} & {0.941} & {0.862}  \\
{UNITE}        & {0.944} & {0.948} & {0.956} & {0.839} \\
{CoCosNet v2}  & {0.959} & {0.955} & {0.963} & {0.877}  \\
{Ours}         & \textbf{0.975} & \textbf{0.961} & \textbf{0.968} & \textbf{0.878}\\
\hline
\end{tabular}}
\caption{Quantitative evaluation of semantic consistency. The higher the score, the better the evaluation metric.}
\label{table:metrics2}
\end{table}

\begin{figure}[t]
\centering
\includegraphics[width=1.0\linewidth]{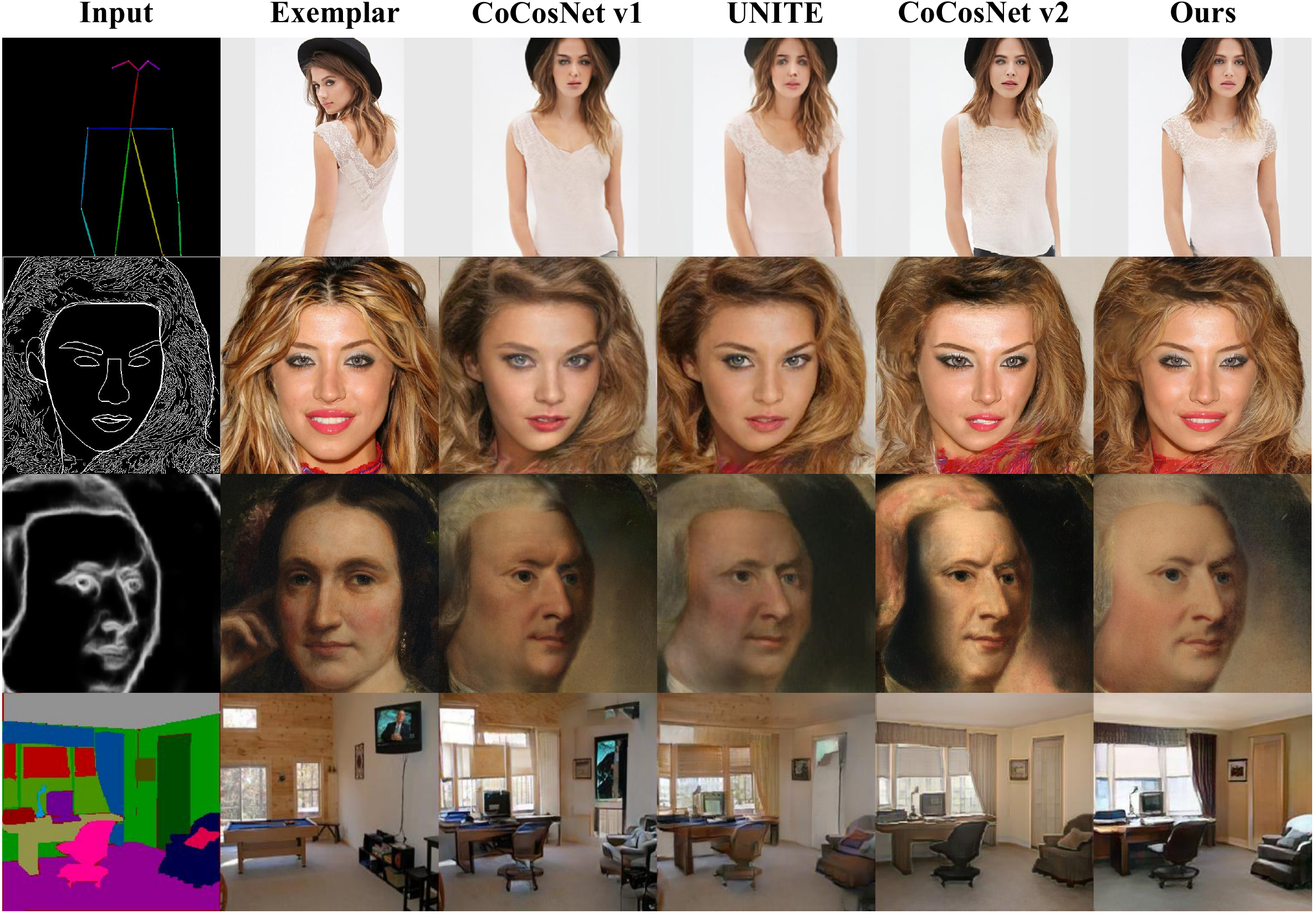}
\caption{Qualitative comparison results of our method with SOTAs on Deepfashion dataset, CelebA-HQ dataset, MetFaces dataset, and ADE20K dataset, respectively.}
\label{fig:compare}
\end{figure}

\begin{table}[t]
\centering
\setlength{\tabcolsep}{0.4mm}{
\begin{tabular}{l|c|c|c|c}
\hline
% \noalign{\smallskip}
& {Deepfashion} & {CelebA-HQ} & {Metfaces} & {ADE20K}\\
% \cline{2-9} 
% & \text { Color } & \text { Texture } & \text { Color } & \text { Texture } & \text { Color } & \text { Texture } & \text { Color } & \text { Texture } \\
% \noalign{\smallskip}
\hline
% \noalign{\smallskip}
{CoCosNet v1}  & {0.959} & {0.947} & {0.944} & {0.952} \\
{UNITE}        & {0.966} & {0.953} & {0.965} & {0.955} \\
{CoCosNet v2}  & {0.974} & {0.966} & {0.964} & \textbf{0.959} \\
{Ours}         & \textbf{0.982} & \textbf{0.974} & \textbf{0.971} & 0.954\\
\hline
\end{tabular}}
\caption{Quantitative evaluation of style similarity. The higher the score, the better the evaluation metric.}
\label{table:metrics3}
\end{table}

\begin{figure}[t]
\centering
\includegraphics[width=1.0\linewidth]{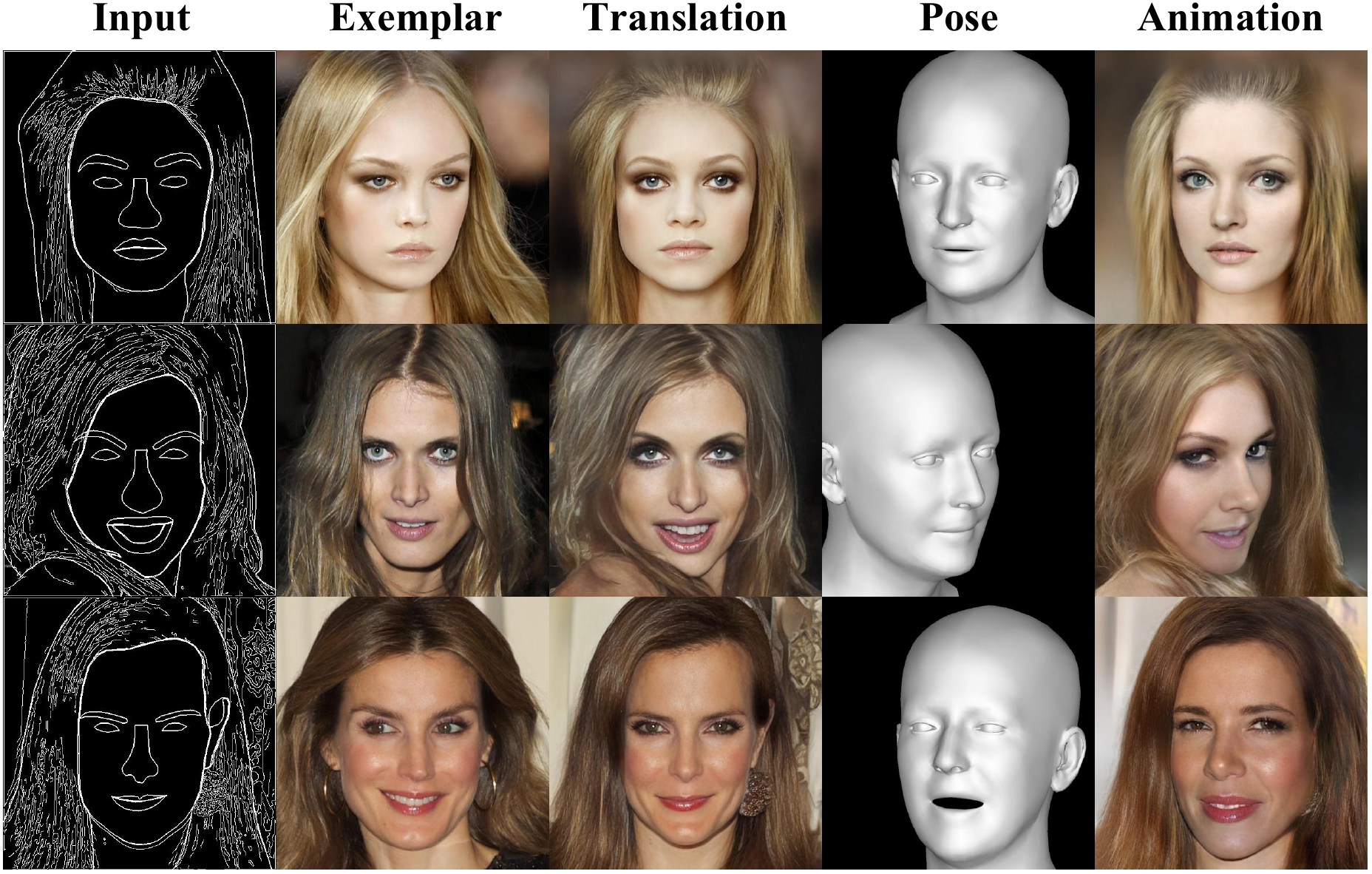}
\caption{Multi-domain image translation results. Through the figure \ref{fig:model2} structure, input edge and exemplar we can get translation result, and then input head pose we can get face animation result based on the translation result.}
\label{fig:animation}
\end{figure}

\begin{figure}[t]
\centering
\includegraphics[width=0.98\linewidth]{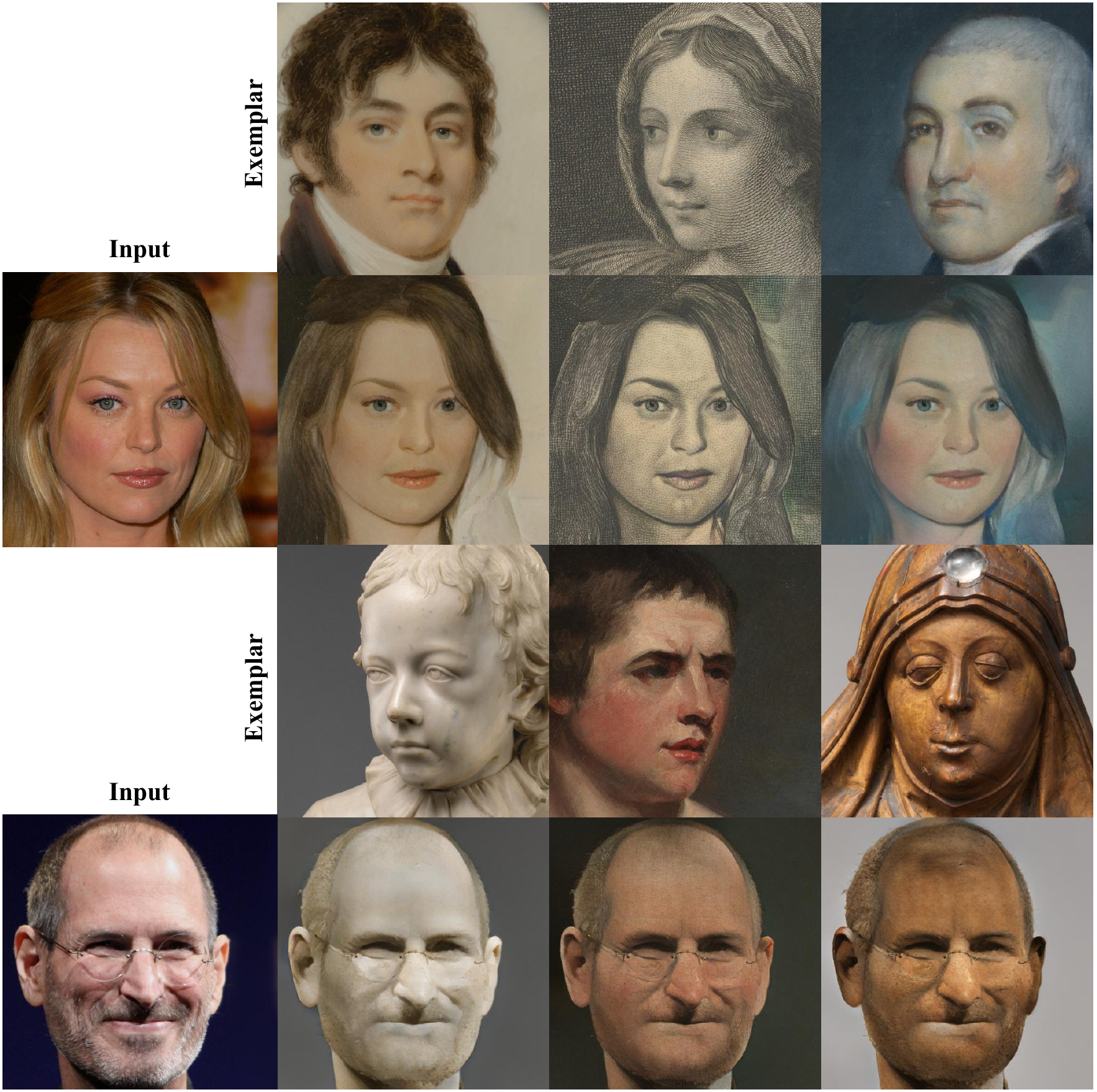}
\caption{Portrait stylization. Our method can transfer image styles from the MetFaces to the faces of the CelebA-HQ. The model is only trained on the
MetFaces dataset. The model input is the edge image of input face.}
\label{fig:cross}
\end{figure}

\begin{figure}[t]
\centering
\includegraphics[width=0.98\linewidth]{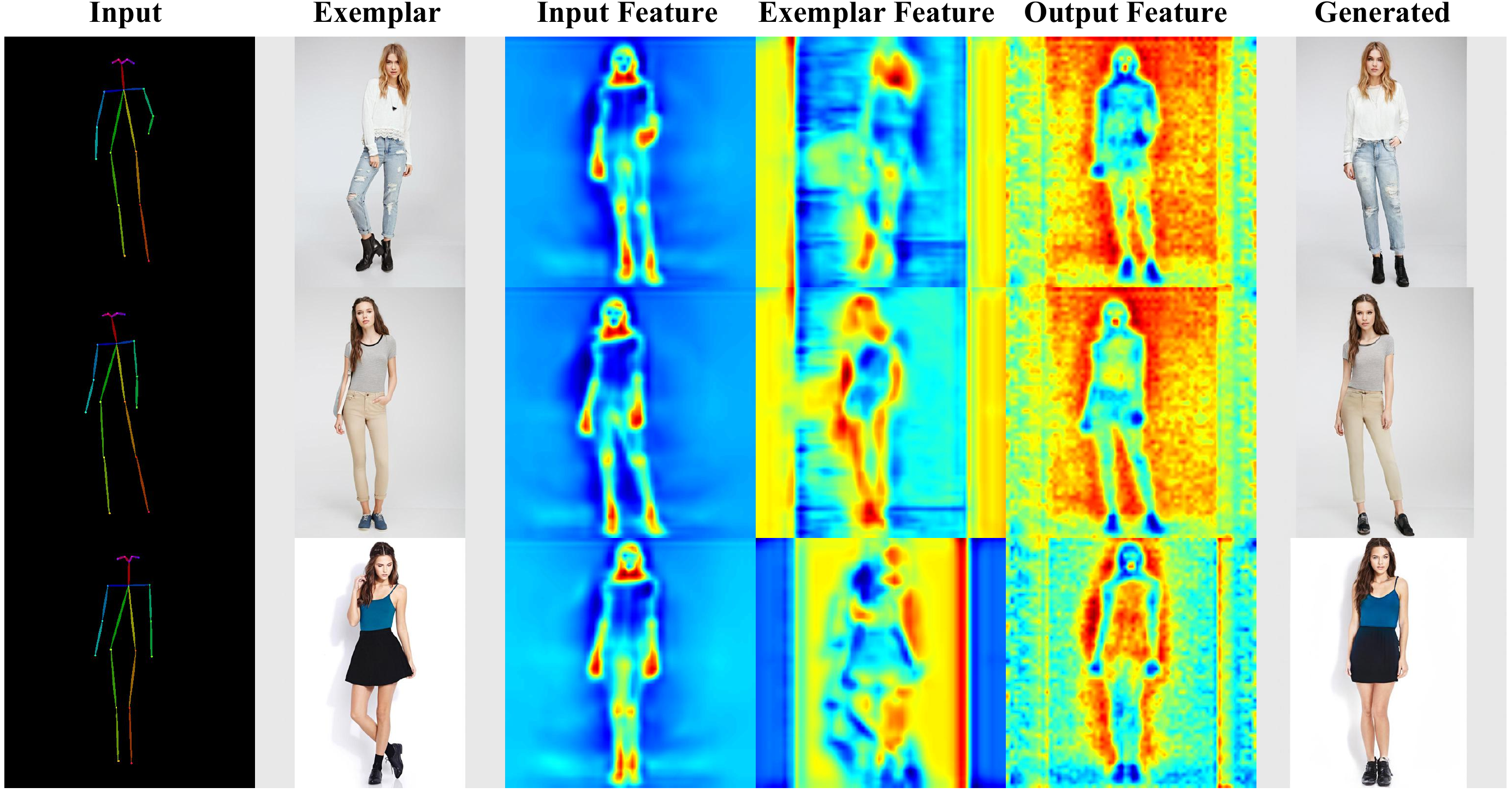}
\caption{Visualization results of CFFT's input and output feature maps, which demonstrate the decoupling and semantic alignment capabilities of CFFT.}
\label{fig:hotmap}
\end{figure}

\noindent \textbf{Implementation Details.}
% 参数设置，训练设置，gpu使用
We use the TTUR \cite{heusel2017gans} learning strategy with a generator and discriminator learning rate of 1e-4 and 4e-4, respectively. We utilize Adam solver with $\beta_1 = 0$ and $\beta_2 = 0.999$. Our experiments are carried out on 8 32GB Tesla V100 GPUs. The input and output feature scales for the CFFT module are $(C,H,W)=(64,64,64)$. The weights of the total loss are $\lambda_{\text{align}}=10$, $\lambda_{\text{match}}=10$, $\lambda_{\text{perc}}=0.001$, $\lambda_{\text{CX}}=10$, $\lambda_{\text{adv}}=10$.

\begin{table}[t]
\centering
\begin{tabular}{l|c|c|c}
\hline
% \noalign{\smallskip}
& {FID} & {SWD} & {Params} \\
\hline
% \noalign{\smallskip}
% {w/o $\mathbf{w}^{+}$} & {} & {} & {x}  \\
{CFFT$\rightarrow$ CM} & {14.3} & {28.8} & {4.8M}  \\
{CFFT$\rightarrow$ AdaIN}  & {16.3} & {30.2} & {0.02M}  \\ 
% {CFFT 16$\times$16}  & {10.3} & {29.0} & {0.88M} \\
{CFFT w/o Hiformer}  & {11.3} & {25.0} & {1.37M} \\
% {CFFT 32$\times$32}  & {9.5} & {23.7} &  {0.98M}  \\
{CFFT 32$\times$32}  & {9.5} & {23.7} &  {2.22M}  \\
{Ours}               & \textbf{8.8} & \textbf{19.0} & 2.62M \\
\hline
\end{tabular}
\caption{Ablation studies of our method. Evaluate the FID and SWD metrics on the Deepfashion dataset, and compute the parameters of each module.}
\label{table:ablation}
\end{table}

\subsection{Experimental Results}

\noindent \textbf{Quantitative Evaluation.} For the quantitative comparison experiments, we first used FID and SWD to assess the quality of the generated images for each model, and use LPIPS to evaluate the model decoupling and semantic alignment ability of each model on the Deepfashion dataset. As shown in Table \ref{table:metrics1}, our approach outperforms the SOTA methods in most datasets. This is due to the fact that our proposed CFFT module is able to better learn the correspondence between different domains and the potentially beneficial information within each domain, allowing the generator to obtain richer information.

In addition to evaluating the quality of the generated images, we quantitatively assess the semantic consistency of the generated images and content inputs and the style similarity to the exemplar images, as shown in Table \ref{table:metrics2} and \ref{table:metrics3}. Compared with the SOTA methods, our model also achieves higher performance.

\noindent \textbf{Qualitative Comparison.} We make a qualitative comparison with SOTA methods, as shown in Figure \ref{fig:compare}. Our method can better fuse the style of the exemplar and the semantic content of the input. Our approach can also obtain more refined results when the semantic difference between exemplar and input is large, as shown in the third row. In addition to the comparative results, Figure \ref{fig:show} also show the more diverse generative results of our method. As can be seen from the figures our approach shows strong performance in a wide range of image translation tasks and the diversity of the generated results reflects the generalizability of our model.

\noindent \textbf{Applications.} Our method has good potential for application. Firstly, we can decouple and fuse information from multiple domains by cascading CFFT modules. We test a specific task, face animation. As shown in Figure \ref{fig:animation}, we first learn exemplar-based image translation between edge input and RGB exemplar, and then add head pose to learn the correspondence between three domains. The head pose is extracted via DECA \cite{feng2021learning}. The whole training process does not use paired data, and it is self-supervised training. Our generated results have the shape, style, and pose corresponding to the each input domain. By cascading CFFT modules, our approach can progressively learn the feature fusion from multiple image domains to achieve more complex image translation. Secondly, we test the portrait stylization experiment, as shown in Figure \ref{fig:cross}. We demonstrate the generalizability and excellent performance of our method through this cross-dataset experiment.

More comparative results and visualizations are presented in the supplementary material.

\subsection{Ablation Study}
We perform the ablation studies of CFFT, the key module of our model, on the Deepfashion dataset. The quantitative results are shown in Table \ref{table:ablation}. The first row of the table represents that we replace the CFFT module with the Correlation Matrix in CoCosNet. This method requires a larger amount of parameters and does not achieve good performance. In the second row of the table we replace the CFFT with the AdaIN \cite{huang2017arbitrary} structure, which is a very common method of feature transfer. Although it has few parameters, its performance is low for complex correspondence learning. In the third row, we remove the Hiformer in the CFFT, and the performance of the model has dropped significantly. In the fourth row, we decrease the spatial scale of the latent features in CFFT, from 64$\times$64 to 32$\times$32. But the model is still able to maintain good performance. In addition, we visualise the input and output features of the CFFT module, as shown in Figure \ref{fig:hotmap}. The results show that the CFFT is capable of good feature decoupling and semantic alignment.

\section{Conclusion}
This paper proposes CFFT-GAN, an exemplar-based image translation that learns the feature fusion of different domains by transformer-based module CFFT, which is able to not only learn the correspondence between domains better but also extract potential beneficial information within each domain. Our proposed CFFT has strong generality, and we can achieve multi-domain image translation by cascading CFFT. Quantitative and qualitative experiments demonstrate the superiority of our approach over the SOTA methods, and our ablation studies on the key module CFFT also reflect its core effect. There is still room for improvement in our approach, e.g., our model still has a performance deficit in the task of mask-to-image. We will optimize this task in future work.

\section{Acknowledgments}
This work is supported by the National Key Research and Development Program of China under Grant No. 2021YFC3320103.

% Use \bibliography{yourbibfile} instead or the References section will not appear in your paper

% \clearpage

\bibliography{aaai23}

\begin{thebibliography}{51}
\providecommand{\natexlab}[1]{#1}

\bibitem[{Arnab et~al.(2021)Arnab, Dehghani, Heigold, Sun, Lu{\v{c}}i{\'c}, and
  Schmid}]{arnab2021vivit}
Arnab, A.; Dehghani, M.; Heigold, G.; Sun, C.; Lu{\v{c}}i{\'c}, M.; and Schmid,
  C. 2021.
\newblock Vivit: A video vision transformer.
\newblock \emph{Proceedings of the IEEE/CVF International Conference on
  Computer Vision}, 6836--6846.

\bibitem[{Barnes et~al.(2009)Barnes, Shechtman, Finkelstein, and
  Goldman}]{barnes2009patchmatch}
Barnes, C.; Shechtman, E.; Finkelstein, A.; and Goldman, D.~B. 2009.
\newblock PatchMatch: A randomized correspondence algorithm for structural
  image editing.
\newblock \emph{ACM Trans. Graph.}, 28(3): 24.

\bibitem[{Cao et~al.(2017)Cao, Simon, Wei, and Sheikh}]{cao2017realtime}
Cao, Z.; Simon, T.; Wei, S.-E.; and Sheikh, Y. 2017.
\newblock Realtime multi-person 2d pose estimation using part affinity fields.
\newblock \emph{Proceedings of the IEEE conference on computer vision and
  pattern recognition}, 7291--7299.

\bibitem[{Carion et~al.(2020)Carion, Massa, Synnaeve, Usunier, Kirillov, and
  Zagoruyko}]{carion2020end}
Carion, N.; Massa, F.; Synnaeve, G.; Usunier, N.; Kirillov, A.; and Zagoruyko,
  S. 2020.
\newblock End-to-end object detection with transformers.
\newblock \emph{European conference on computer vision}, 213--229.

\bibitem[{Choi et~al.(2018)Choi, Choi, Kim, Ha, Kim, and
  Choo}]{choi2018stargan}
Choi, Y.; Choi, M.; Kim, M.; Ha, J.-W.; Kim, S.; and Choo, J. 2018.
\newblock Stargan: Unified generative adversarial networks for multi-domain
  image-to-image translation.
\newblock \emph{Proceedings of the IEEE conference on computer vision and
  pattern recognition}, 8789--8797.

\bibitem[{Choi et~al.(2020)Choi, Uh, Yoo, and Ha}]{choi2020stargan}
Choi, Y.; Uh, Y.; Yoo, J.; and Ha, J.-W. 2020.
\newblock Stargan v2: Diverse image synthesis for multiple domains.
\newblock \emph{Proceedings of the IEEE/CVF conference on computer vision and
  pattern recognition}, 8188--8197.

\bibitem[{Dosovitskiy et~al.(2021)Dosovitskiy, Beyer, Kolesnikov, Weissenborn,
  Zhai, Unterthiner, Dehghani, Minderer, Heigold, Gelly, Uszkoreit, and
  Houlsby}]{dosovitskiy2020vit}
Dosovitskiy, A.; Beyer, L.; Kolesnikov, A.; Weissenborn, D.; Zhai, X.;
  Unterthiner, T.; Dehghani, M.; Minderer, M.; Heigold, G.; Gelly, S.;
  Uszkoreit, J.; and Houlsby, N. 2021.
\newblock An Image is Worth 16x16 Words: Transformers for Image Recognition at
  Scale.
\newblock \emph{ICLR}.

\bibitem[{Feng et~al.(2021)Feng, Feng, Black, and Bolkart}]{feng2021learning}
Feng, Y.; Feng, H.; Black, M.~J.; and Bolkart, T. 2021.
\newblock Learning an animatable detailed 3D face model from in-the-wild
  images.
\newblock \emph{ACM Transactions on Graphics (TOG)}, 40(4): 1--13.

\bibitem[{Heusel et~al.(2017)Heusel, Ramsauer, Unterthiner, Nessler, and
  Hochreiter}]{heusel2017gans}
Heusel, M.; Ramsauer, H.; Unterthiner, T.; Nessler, B.; and Hochreiter, S.
  2017.
\newblock Gans trained by a two time-scale update rule converge to a local nash
  equilibrium.
\newblock \emph{Advances in neural information processing systems}, 30.

\bibitem[{Huang and Belongie(2017)}]{huang2017arbitrary}
Huang, X.; and Belongie, S. 2017.
\newblock Arbitrary style transfer in real-time with adaptive instance
  normalization.
\newblock \emph{Proceedings of the IEEE international conference on computer
  vision}, 1501--1510.

\bibitem[{Huang et~al.(2018)Huang, Liu, Belongie, and
  Kautz}]{huang2018multimodal}
Huang, X.; Liu, M.-Y.; Belongie, S.; and Kautz, J. 2018.
\newblock Multimodal unsupervised image-to-image translation.
\newblock \emph{Proceedings of the European conference on computer vision
  (ECCV)}, 172--189.

\bibitem[{Isola et~al.(2017)Isola, Zhu, Zhou, and Efros}]{isola2017image}
Isola, P.; Zhu, J.-Y.; Zhou, T.; and Efros, A.~A. 2017.
\newblock Image-to-image translation with conditional adversarial networks.
\newblock \emph{Proceedings of the IEEE conference on computer vision and
  pattern recognition}, 1125--1134.

\bibitem[{Jiang, Chang, and Wang(2021)}]{jiang2021transgan}
Jiang, Y.; Chang, S.; and Wang, Z. 2021.
\newblock Transgan: Two pure transformers can make one strong gan, and that can
  scale up.
\newblock \emph{Advances in Neural Information Processing Systems}, 34.

\bibitem[{Johnson, Alahi, and Fei-Fei(2016)}]{johnson2016perceptual}
Johnson, J.; Alahi, A.; and Fei-Fei, L. 2016.
\newblock Perceptual losses for real-time style transfer and super-resolution.
\newblock \emph{European conference on computer vision}, 694--711.

\bibitem[{Karras et~al.(2017)Karras, Aila, Laine, and
  Lehtinen}]{karras2017progressive}
Karras, T.; Aila, T.; Laine, S.; and Lehtinen, J. 2017.
\newblock Progressive growing of gans for improved quality, stability, and
  variation.
\newblock \emph{arXiv preprint arXiv:1710.10196}.

\bibitem[{Karras et~al.(2020)Karras, Aittala, Hellsten, Laine, Lehtinen, and
  Aila}]{karras2020training}
Karras, T.; Aittala, M.; Hellsten, J.; Laine, S.; Lehtinen, J.; and Aila, T.
  2020.
\newblock Training generative adversarial networks with limited data.
\newblock \emph{Advances in Neural Information Processing Systems}, 33:
  12104--12114.

\bibitem[{Lee et~al.(2018)Lee, Tseng, Huang, Singh, and Yang}]{lee2018diverse}
Lee, H.-Y.; Tseng, H.-Y.; Huang, J.-B.; Singh, M.; and Yang, M.-H. 2018.
\newblock Diverse image-to-image translation via disentangled representations.
\newblock \emph{Proceedings of the European conference on computer vision
  (ECCV)}, 35--51.

\bibitem[{Lee et~al.(2021)Lee, Chang, Jiang, Zhang, Tu, and
  Liu}]{lee2021vitgan}
Lee, K.; Chang, H.; Jiang, L.; Zhang, H.; Tu, Z.; and Liu, C. 2021.
\newblock Vitgan: Training gans with vision transformers.
\newblock \emph{arXiv preprint arXiv:2107.04589}.

\bibitem[{Li et~al.(2020)Li, Jiao, Cao, Wong, and Wu}]{li2020model}
Li, R.; Jiao, Q.; Cao, W.; Wong, H.-S.; and Wu, S. 2020.
\newblock Model adaptation: Unsupervised domain adaptation without source data.
\newblock \emph{Proceedings of the IEEE/CVF Conference on Computer Vision and
  Pattern Recognition}, 9641--9650.

\bibitem[{Li, Huang, and Loy(2019)}]{li2019dense}
Li, Y.; Huang, C.; and Loy, C.~C. 2019.
\newblock Dense intrinsic appearance flow for human pose transfer.
\newblock \emph{Proceedings of the IEEE/CVF Conference on Computer Vision and
  Pattern Recognition}, 3693--3702.

\bibitem[{Liu et~al.(2022)Liu, Ye, Ren, and Wang}]{liu2022dynast}
Liu, S.; Ye, J.; Ren, S.; and Wang, X. 2022.
\newblock Dynast: Dynamic sparse transformer for exemplar-guided image
  generation.
\newblock In \emph{Computer Vision--ECCV 2022: 17th European Conference, Tel
  Aviv, Israel, October 23--27, 2022, Proceedings, Part XVI}, 72--90. Springer.

\bibitem[{Liu, Zhang, and Wang(2021)}]{liu2021source}
Liu, Y.; Zhang, W.; and Wang, J. 2021.
\newblock Source-free domain adaptation for semantic segmentation.
\newblock \emph{Proceedings of the IEEE/CVF Conference on Computer Vision and
  Pattern Recognition}, 1215--1224.

\bibitem[{Liu et~al.(2021)Liu, Lin, Cao, Hu, Wei, Zhang, Lin, and
  Guo}]{liu2021swin}
Liu, Z.; Lin, Y.; Cao, Y.; Hu, H.; Wei, Y.; Zhang, Z.; Lin, S.; and Guo, B.
  2021.
\newblock Swin transformer: Hierarchical vision transformer using shifted
  windows.
\newblock \emph{Proceedings of the IEEE/CVF International Conference on
  Computer Vision}, 10012--10022.

\bibitem[{Liu et~al.(2016)Liu, Luo, Qiu, Wang, and Tang}]{liu2016deepfashion}
Liu, Z.; Luo, P.; Qiu, S.; Wang, X.; and Tang, X. 2016.
\newblock Deepfashion: Powering robust clothes recognition and retrieval with
  rich annotations.
\newblock \emph{Proceedings of the IEEE conference on computer vision and
  pattern recognition}, 1096--1104.

\bibitem[{Liu et~al.(2015)Liu, Luo, Wang, and Tang}]{liu2015deep}
Liu, Z.; Luo, P.; Wang, X.; and Tang, X. 2015.
\newblock Deep learning face attributes in the wild.
\newblock \emph{Proceedings of the IEEE international conference on computer
  vision}, 3730--3738.

\bibitem[{Ma et~al.(2018)Ma, Jia, Georgoulis, Tuytelaars, and
  Van~Gool}]{ma2018exemplar}
Ma, L.; Jia, X.; Georgoulis, S.; Tuytelaars, T.; and Van~Gool, L. 2018.
\newblock Exemplar guided unsupervised image-to-image translation with semantic
  consistency.
\newblock \emph{arXiv preprint arXiv:1805.11145}.

\bibitem[{Ma et~al.(2017)Ma, Jia, Sun, Schiele, Tuytelaars, and
  Van~Gool}]{ma2017pose}
Ma, L.; Jia, X.; Sun, Q.; Schiele, B.; Tuytelaars, T.; and Van~Gool, L. 2017.
\newblock Pose guided person image generation.
\newblock \emph{Advances in neural information processing systems}, 30.

\bibitem[{Ma et~al.(2021)Ma, Peng, Wang, and Dong}]{ma2021must}
Ma, T.; Peng, B.; Wang, W.; and Dong, J. 2021.
\newblock MUST-GAN: Multi-level Statistics Transfer for Self-driven Person
  Image Generation.
\newblock \emph{Proceedings of the IEEE/CVF Conference on Computer Vision and
  Pattern Recognition}, 13622--13631.

\bibitem[{Mechrez, Talmi, and Zelnik-Manor(2018)}]{mechrez2018contextual}
Mechrez, R.; Talmi, I.; and Zelnik-Manor, L. 2018.
\newblock The contextual loss for image transformation with non-aligned data.
\newblock \emph{Proceedings of the European conference on computer vision
  (ECCV)}, 768--783.

\bibitem[{Men et~al.(2020)Men, Mao, Jiang, Ma, and Lian}]{men2020controllable}
Men, Y.; Mao, Y.; Jiang, Y.; Ma, W.-Y.; and Lian, Z. 2020.
\newblock Controllable person image synthesis with attribute-decomposed gan.
\newblock \emph{Proceedings of the IEEE/CVF Conference on Computer Vision and
  Pattern Recognition}, 5084--5093.

\bibitem[{Park et~al.(2020)Park, Efros, Zhang, and Zhu}]{park2020contrastive}
Park, T.; Efros, A.~A.; Zhang, R.; and Zhu, J.-Y. 2020.
\newblock Contrastive learning for unpaired image-to-image translation.
\newblock \emph{European Conference on Computer Vision}, 319--345.

\bibitem[{Park et~al.(2019)Park, Liu, Wang, and Zhu}]{park2019semantic}
Park, T.; Liu, M.-Y.; Wang, T.-C.; and Zhu, J.-Y. 2019.
\newblock Semantic image synthesis with spatially-adaptive normalization.
\newblock \emph{Proceedings of the IEEE/CVF conference on computer vision and
  pattern recognition}, 2337--2346.

\bibitem[{Ren et~al.(2020)Ren, Yu, Chen, Li, and Li}]{ren2020deep}
Ren, Y.; Yu, X.; Chen, J.; Li, T.~H.; and Li, G. 2020.
\newblock Deep image spatial transformation for person image generation.
\newblock \emph{Proceedings of the IEEE/CVF Conference on Computer Vision and
  Pattern Recognition}, 7690--7699.

\bibitem[{Saito, Saenko, and Liu(2020)}]{saito2020coco}
Saito, K.; Saenko, K.; and Liu, M.-Y. 2020.
\newblock Coco-funit: Few-shot unsupervised image translation with a content
  conditioned style encoder.
\newblock \emph{European Conference on Computer Vision}, 382--398.

\bibitem[{Shen et~al.(2020)Shen, Gu, Tang, and Zhou}]{shen2020interpreting}
Shen, Y.; Gu, J.; Tang, X.; and Zhou, B. 2020.
\newblock Interpreting the latent space of gans for semantic face editing.
\newblock \emph{Proceedings of the IEEE/CVF Conference on Computer Vision and
  Pattern Recognition}, 9243--9252.

\bibitem[{Siarohin et~al.(2018)Siarohin, Sangineto, Lathuiliere, and
  Sebe}]{siarohin2018deformable}
Siarohin, A.; Sangineto, E.; Lathuiliere, S.; and Sebe, N. 2018.
\newblock Deformable gans for pose-based human image generation.
\newblock \emph{Proceedings of the IEEE Conference on Computer Vision and
  Pattern Recognition}, 3408--3416.

\bibitem[{Vaswani et~al.(2017)Vaswani, Shazeer, Parmar, Uszkoreit, Jones,
  Gomez, Kaiser, and Polosukhin}]{vaswani2017attention}
Vaswani, A.; Shazeer, N.; Parmar, N.; Uszkoreit, J.; Jones, L.; Gomez, A.~N.;
  Kaiser, {\L}.; and Polosukhin, I. 2017.
\newblock Attention is all you need.
\newblock \emph{Advances in neural information processing systems}, 30.

\bibitem[{Wang et~al.(2020)Wang, Zhu, Green, Adam, Yuille, and
  Chen}]{wang2020axial}
Wang, H.; Zhu, Y.; Green, B.; Adam, H.; Yuille, A.; and Chen, L.-C. 2020.
\newblock Axial-deeplab: Stand-alone axial-attention for panoptic segmentation.
\newblock \emph{European Conference on Computer Vision}, 108--126.

\bibitem[{Wang et~al.(2019)Wang, Yang, Li, Liang, Zhang, Hall, and
  Hu}]{wang2019example}
Wang, M.; Yang, G.-Y.; Li, R.; Liang, R.-Z.; Zhang, S.-H.; Hall, P.~M.; and Hu,
  S.-M. 2019.
\newblock Example-guided style-consistent image synthesis from semantic
  labeling.
\newblock \emph{Proceedings of the IEEE/CVF Conference on Computer Vision and
  Pattern Recognition}, 1495--1504.

\bibitem[{Wang et~al.(2018)Wang, Liu, Zhu, Tao, Kautz, and
  Catanzaro}]{wang2018high}
Wang, T.-C.; Liu, M.-Y.; Zhu, J.-Y.; Tao, A.; Kautz, J.; and Catanzaro, B.
  2018.
\newblock High-resolution image synthesis and semantic manipulation with
  conditional gans.
\newblock \emph{Proceedings of the IEEE conference on computer vision and
  pattern recognition}, 8798--8807.

\bibitem[{Wu et~al.(2020)Wu, Zhang, Lu, and Chen}]{wu2020cascade}
Wu, R.; Zhang, G.; Lu, S.; and Chen, T. 2020.
\newblock Cascade ef-gan: Progressive facial expression editing with local
  focuses.
\newblock \emph{Proceedings of the IEEE/CVF Conference on Computer Vision and
  Pattern Recognition}, 5021--5030.

\bibitem[{Xie and Tu(2015)}]{xie2015holistically}
Xie, S.; and Tu, Z. 2015.
\newblock Holistically-nested edge detection.
\newblock \emph{Proceedings of the IEEE international conference on computer
  vision}, 1395--1403.

\bibitem[{Yi et~al.(2017)Yi, Zhang, Tan, and Gong}]{yi2017dualgan}
Yi, Z.; Zhang, H.; Tan, P.; and Gong, M. 2017.
\newblock Dualgan: Unsupervised dual learning for image-to-image translation.
\newblock \emph{Proceedings of the IEEE international conference on computer
  vision}, 2849--2857.

\bibitem[{Yu et~al.(2018)Yu, Lin, Yang, Shen, Lu, and Huang}]{yu2018generative}
Yu, J.; Lin, Z.; Yang, J.; Shen, X.; Lu, X.; and Huang, T.~S. 2018.
\newblock Generative image inpainting with contextual attention.
\newblock \emph{Proceedings of the IEEE conference on computer vision and
  pattern recognition}, 5505--5514.

\bibitem[{Zhan et~al.(2021)Zhan, Yu, Cui, Zhang, Lu, Pan, Zhang, Ma, Xie, and
  Miao}]{zhan2021unbalanced}
Zhan, F.; Yu, Y.; Cui, K.; Zhang, G.; Lu, S.; Pan, J.; Zhang, C.; Ma, F.; Xie,
  X.; and Miao, C. 2021.
\newblock Unbalanced feature transport for exemplar-based image translation.
\newblock \emph{Proceedings of the IEEE/CVF Conference on Computer Vision and
  Pattern Recognition}, 15028--15038.

\bibitem[{Zhang et~al.(2020)Zhang, Zhang, Chen, Yuan, and Wen}]{zhang2020cross}
Zhang, P.; Zhang, B.; Chen, D.; Yuan, L.; and Wen, F. 2020.
\newblock Cross-domain correspondence learning for exemplar-based image
  translation.
\newblock \emph{Proceedings of the IEEE/CVF Conference on Computer Vision and
  Pattern Recognition}, 5143--5153.

\bibitem[{Zhou et~al.(2017)Zhou, Zhao, Puig, Fidler, Barriuso, and
  Torralba}]{zhou2017scene}
Zhou, B.; Zhao, H.; Puig, X.; Fidler, S.; Barriuso, A.; and Torralba, A. 2017.
\newblock Scene parsing through ade20k dataset.
\newblock \emph{Proceedings of the IEEE conference on computer vision and
  pattern recognition}, 633--641.

\bibitem[{Zhou et~al.(2021)Zhou, Zhang, Zhang, Zhang, Bao, Chen, Zhang, and
  Wen}]{zhou2021cocosnet}
Zhou, X.; Zhang, B.; Zhang, T.; Zhang, P.; Bao, J.; Chen, D.; Zhang, Z.; and
  Wen, F. 2021.
\newblock Cocosnet v2: Full-resolution correspondence learning for image
  translation.
\newblock \emph{Proceedings of the IEEE/CVF Conference on Computer Vision and
  Pattern Recognition}, 11465--11475.

\bibitem[{Zhu et~al.(2017{\natexlab{a}})Zhu, Park, Isola, and
  Efros}]{zhu2017unpaired}
Zhu, J.-Y.; Park, T.; Isola, P.; and Efros, A.~A. 2017{\natexlab{a}}.
\newblock Unpaired image-to-image translation using cycle-consistent
  adversarial networks.
\newblock \emph{Proceedings of the IEEE international conference on computer
  vision}, 2223--2232.

\bibitem[{Zhu et~al.(2017{\natexlab{b}})Zhu, Zhang, Pathak, Darrell, Efros,
  Wang, and Shechtman}]{zhu2017toward}
Zhu, J.-Y.; Zhang, R.; Pathak, D.; Darrell, T.; Efros, A.~A.; Wang, O.; and
  Shechtman, E. 2017{\natexlab{b}}.
\newblock Toward multimodal image-to-image translation.
\newblock \emph{Advances in neural information processing systems}, 30.

\bibitem[{Zhu et~al.(2020)Zhu, Abdal, Qin, and Wonka}]{zhu2020sean}
Zhu, P.; Abdal, R.; Qin, Y.; and Wonka, P. 2020.
\newblock Sean: Image synthesis with semantic region-adaptive normalization.
\newblock In \emph{Proceedings of the IEEE/CVF Conference on Computer Vision
  and Pattern Recognition}, 5104--5113.

\end{thebibliography}

\appendix
\newpage
\onecolumn

\section{Implementation Details}
Table \ref{table:structure} shows the detailed architecture of Encoder and Generator. The output feature of CFFT and the scaled input image are concatenated and input to each SPADE ResBlock layer of the Generator.

\begin{table}[ht]
\centering
\begin{tabular}{c|c|c}
\hline
Module                       & Layers in the module          & Output size  \\
\hline
\multirow{9}{*}{Encoder $\times$ 2} & Conv2d / k3s1  & 512$\times$512$\times$64   \\
        & Conv2d / k4s2         & 256$\times$256$\times$128  \\
        & Conv2d / k3s1         & 256$\times$256$\times$256  \\
        & Conv2d / k4s2         & 128$\times$128$\times$256  \\
        & Conv2d / k3s1         & 128$\times$128$\times$512  \\
        & Conv2d / k4s2         & 64$\times$64$\times$512    \\
        & Resblock / k3s1       & 64$\times$64$\times$512    \\
        & Resblock / k3s1       & 64$\times$64$\times$256    \\
        & Resblock / k3s1       & 64$\times$64$\times$64     \\
% \hline
% \multirow{5}{*}{CFFT}     & Flattening, Channel Concat  & 4096$\times$128 \\
%                          & (Resblock: LN, MSA, MLP) $\times$ 3  & 4096$\times$128    \\
%                          & MLP                           & 4096$\times$64      \\
%                          & (Resblock: LN, MSA, MLP) $\times$ 3  & 4096$\times$64      \\
%                          & Reshape                       & 64$\times$64$\times$64     \\
% \hline
% \multirow{3}{*}{$\mathbf{w+}$ layers}   & Upsampling, (Conv2d/k3s1) $\times$ 2 & 128$\times$128$\times$64   \\
%                                 & Upsampling, (Conv2d/k3s1) $\times$ 2 & 256$\times$256$\times$64   \\
%                                 & Upsampling, (Conv2d/k3s1) $\times$ 2 & 512$\times$512$\times$64   \\
\hline
\multirow{9}{*}{Generator}   & Conv2d / k3s1  & 16$\times$16$\times$1024   \\
            & SPADE ResBlock, Upsampling & 32$\times$32$\times$1024   \\
            & SPADE ResBlock & 32$\times$32$\times$1024 \\
            & SPADE ResBlock, Upsampling & 64$\times$64$\times$1024 \\
            & SPADE ResBlock, Upsampling & 128$\times$128$\times$512\\
            & SPADE ResBlock, Upsampling & 256$\times$256$\times$256 \\
            & SPADE ResBlock, Upsampling & 512$\times$512$\times$128 \\
            & SPADE ResBlock & 512$\times$512$\times$64 \\
            & Conv2d / k3s1, Tanh  & 512$\times$512$\times$3   \\

\hline
\end{tabular}
\caption{The detailed architecture of Encoder and Generator. k3s1 indicates the convolutional layer with kernel size 3 and stride 1. }
\label{table:structure}
\end{table}

\clearpage

\section{Additional Experimental Results}
To demonstrate the excellent performance of our proposed CFFT-GAN, we present many additional visualization results as follows. All the experiments are performed on 512$\times$512 resolution.

\begin{figure}[h]
\centering
\includegraphics[width=0.8\textwidth]{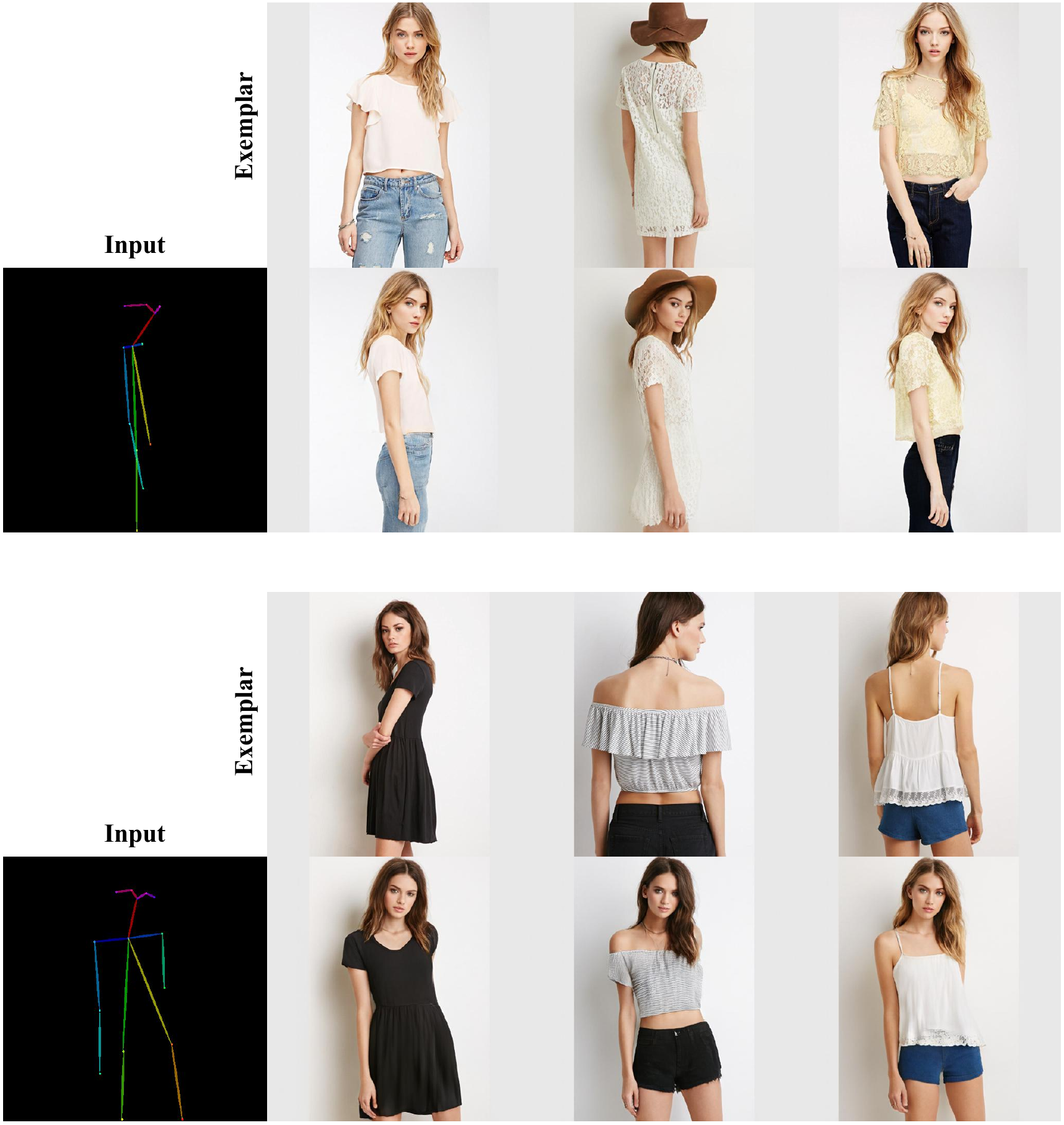}
\caption{Exemplar-based image translation results on Deepfashion dataset at the resolution of 512$\times$512.}
\label{fig:d1}
\end{figure}

\begin{figure}[t]
\centering
\includegraphics[width=0.9\textwidth]{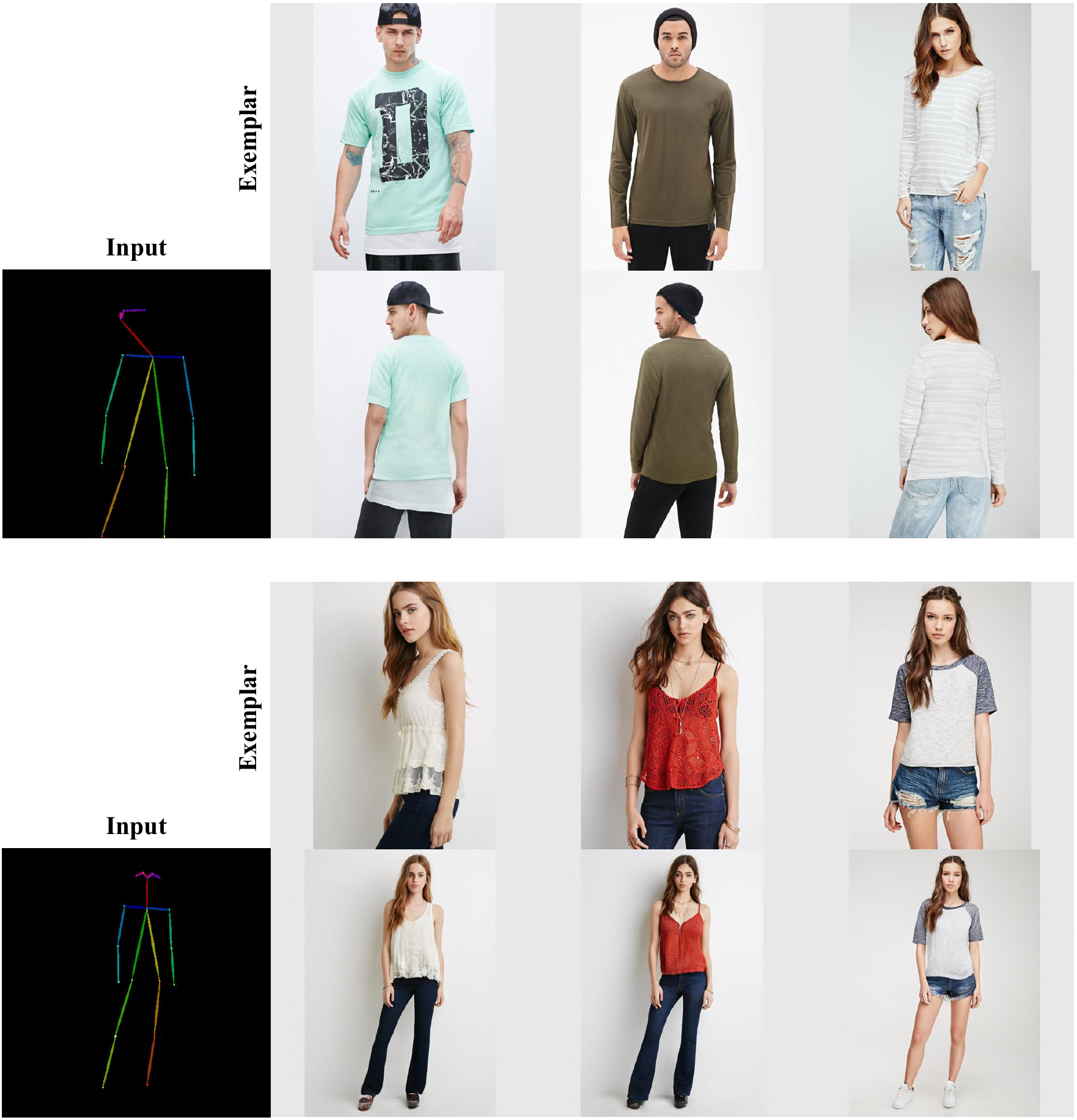}
\caption{Exemplar-based image translation results on Deepfashion dataset at the resolution of 512$\times$512.}
\label{fig:d2}
\end{figure}

\begin{figure}[t]
\centering
\includegraphics[width=0.9\textwidth]{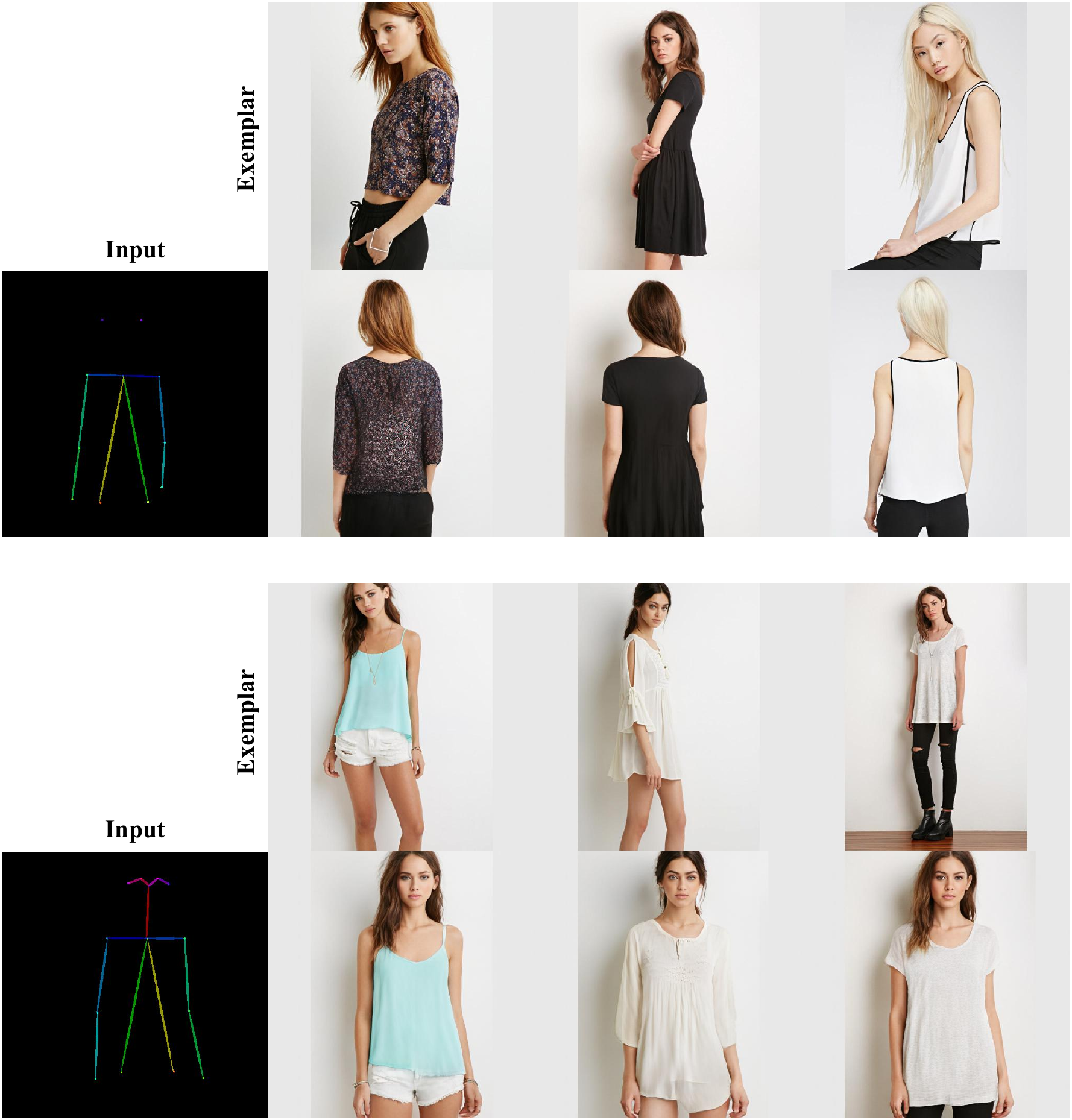}
\caption{Exemplar-based image translation results on Deepfashion dataset at the resolution of 512$\times$512.}
\label{fig:d3}
\end{figure}

Figure \ref{fig:d1}, \ref{fig:d2} and \ref{fig:d3} show more experimental results on the Deepfashion dataset. A wide range of clothing styles are maintained in the image translation process and the identity and details of the faces are maintained very well.

\clearpage

\begin{figure}[h]
\centering
\includegraphics[width=0.9\textwidth]{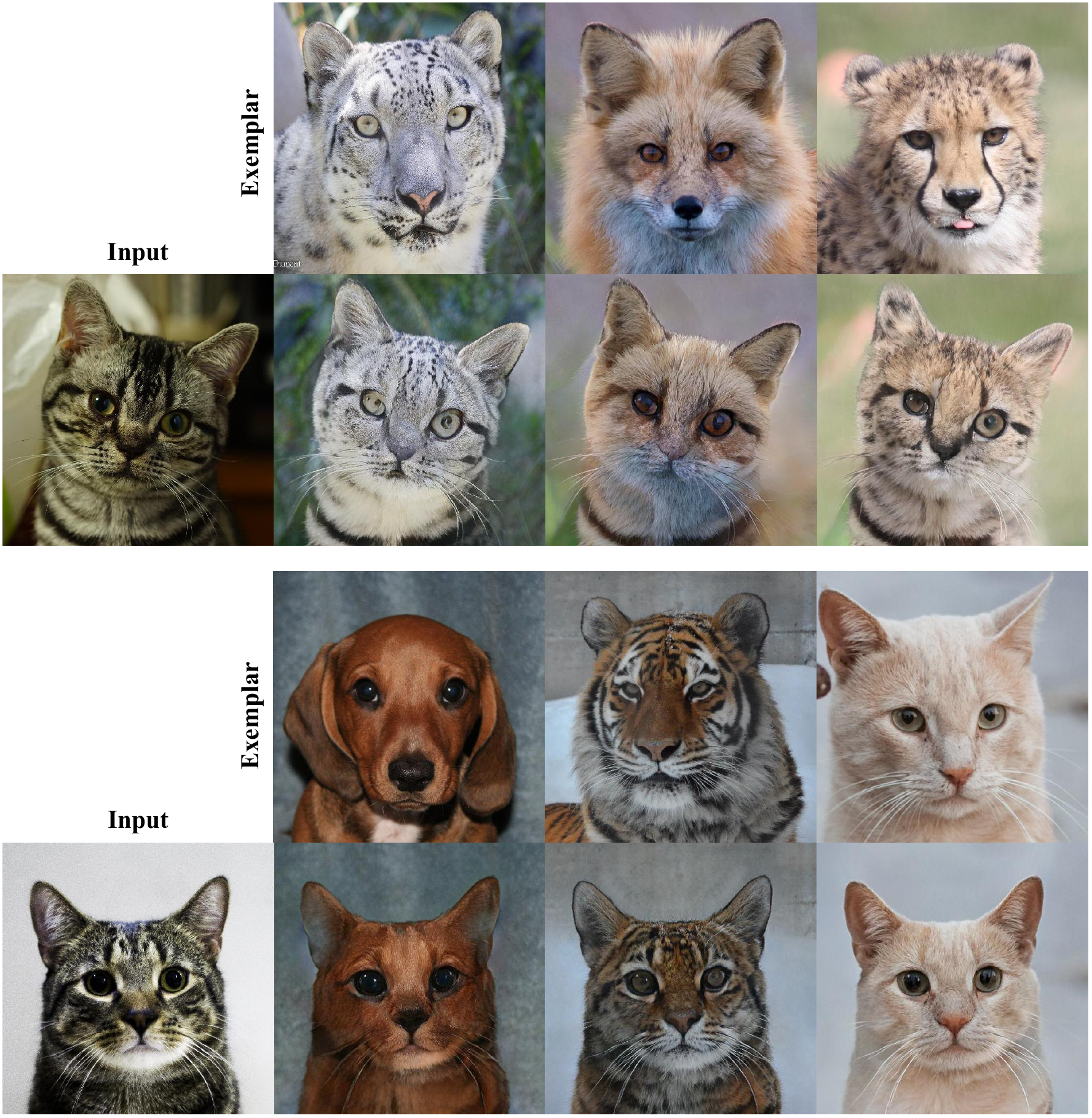}
\caption{Exemplar-based image translation results on AFHQ dataset at the resolution 512$\times$512. The input of the model is the edge of input animal face.}
\label{fig:a1}
\end{figure}

\begin{figure}[t]
\centering
\includegraphics[width=0.9\textwidth]{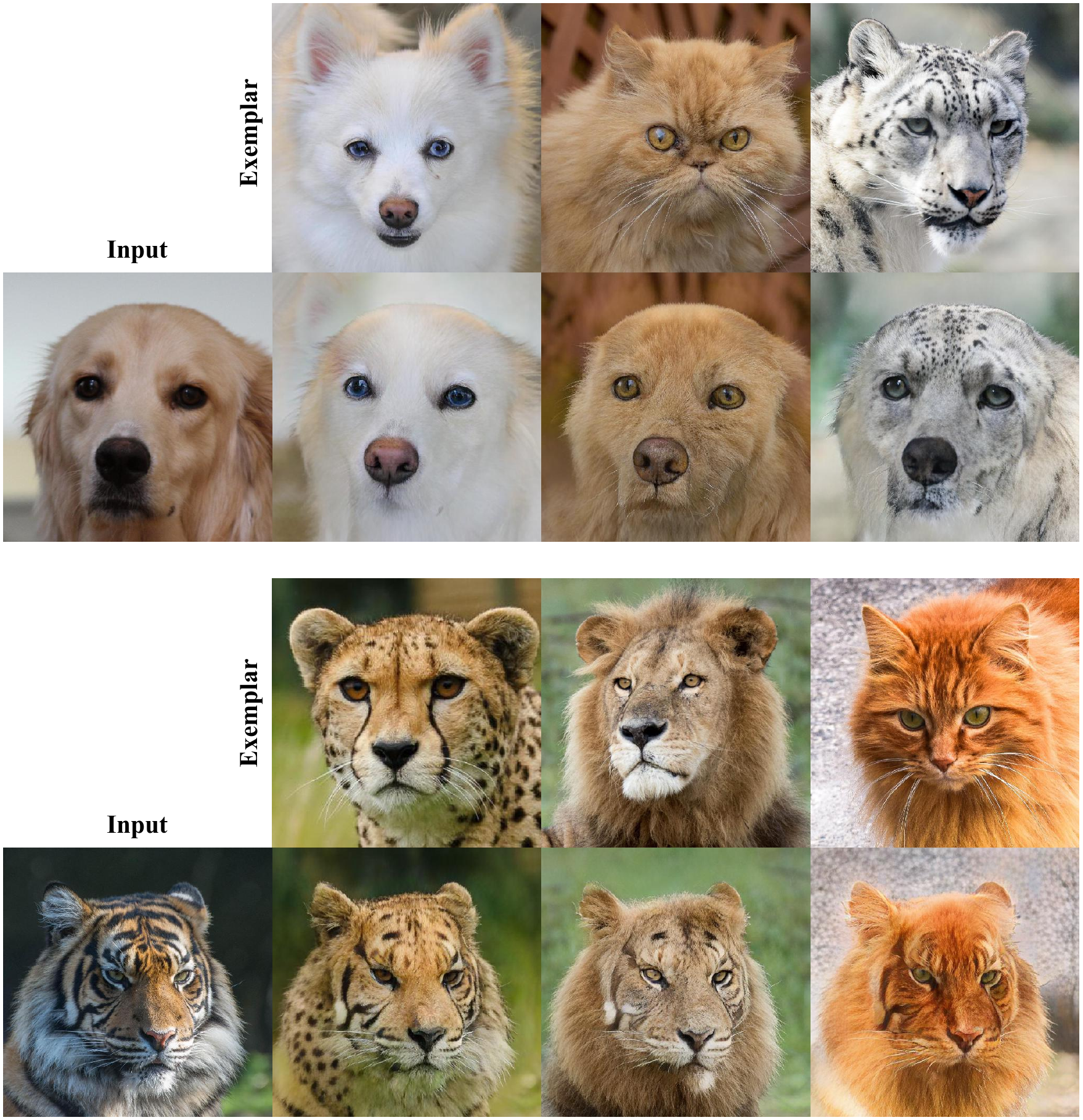}
\caption{Exemplar-based image translation results on AFHQ dataset at the resolution 512$\times$512. The input of the model is the edge of input animal face.}
\label{fig:a2}
\end{figure}

Figure \ref{fig:a1} and \ref{fig:a2} show more image translation results on the AFHQ dataset. Our method is very effective in transferring the hair color and texture of the exemplar to the target animal face and generating realistic animal faces.

\clearpage

\begin{figure}[ht]
\centering
\includegraphics[width=1.0\textwidth]{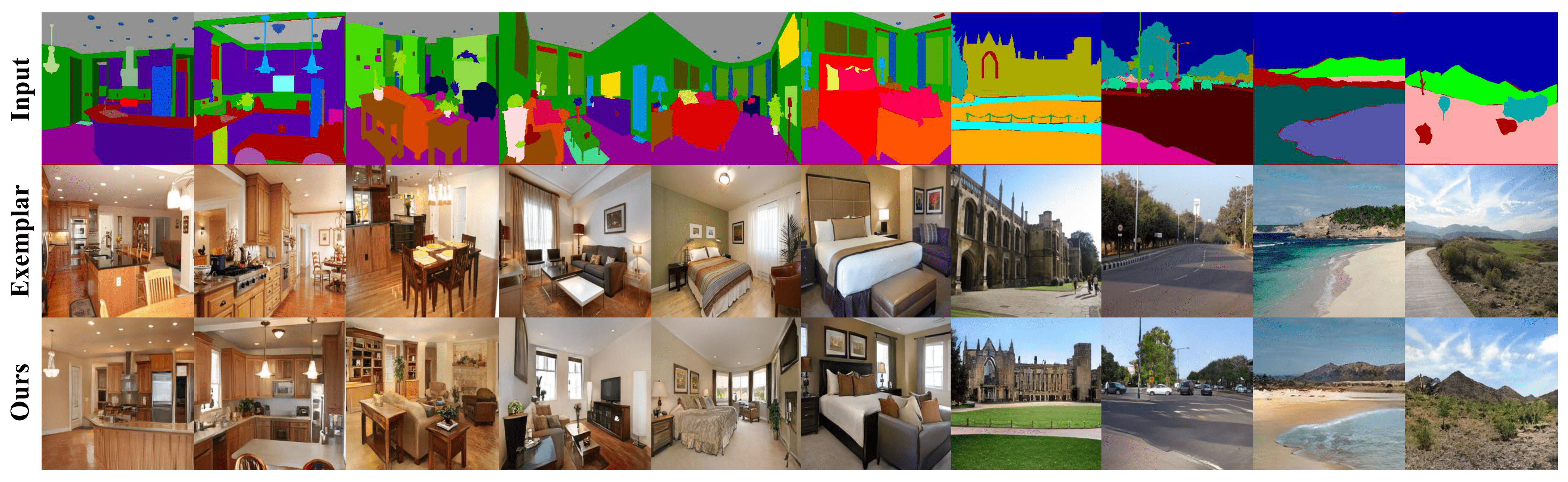}
\caption{Exemplar-based image translation results on ADE20K dataset at the resolution 256$\times$256.}
\label{fig:more_ADE20K}
\end{figure}

Figure \ref{fig:more_ADE20K} shows more image translation results on the ADE20K dataset. Our method can transfer the style and texture of the exemplar image well with ensuring the accurate generation of the input semantic content.

\clearpage

\begin{figure}[h]
\centering
\includegraphics[width=0.9\textwidth]{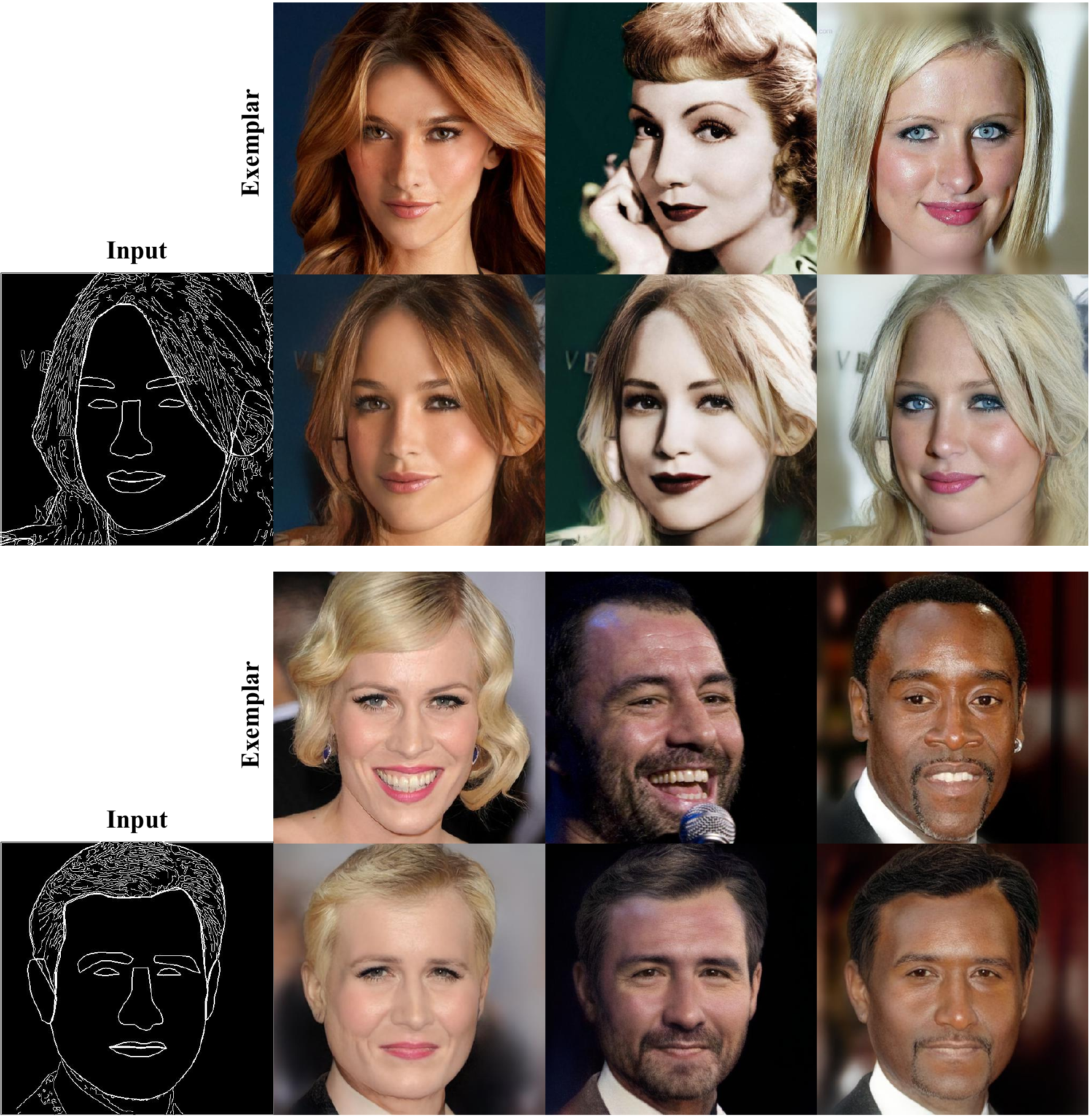}
\caption{Exemplar-based image translation results on CelebA-HQ dataset at the resolution 512$\times$512. }
\label{fig:ce1}
\end{figure}

\begin{figure}[t]
\centering
\includegraphics[width=0.9\textwidth]{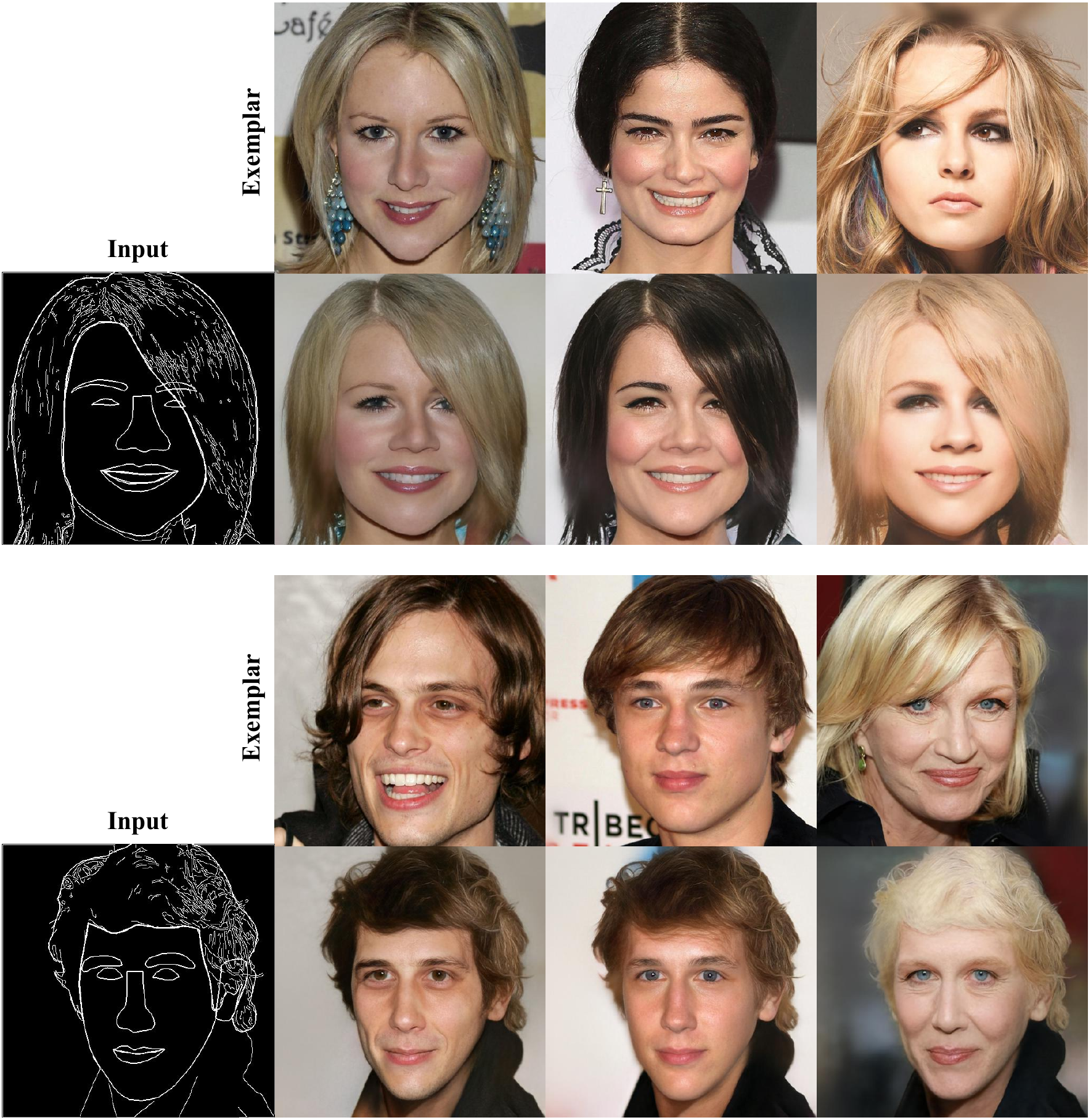}
\caption{Exemplar-based image translation results on CelebA-HQ dataset at the resolution 512$\times$512. }
\label{fig:ce2}
\end{figure}

\begin{figure}[t]
\centering
\includegraphics[width=0.9\textwidth]{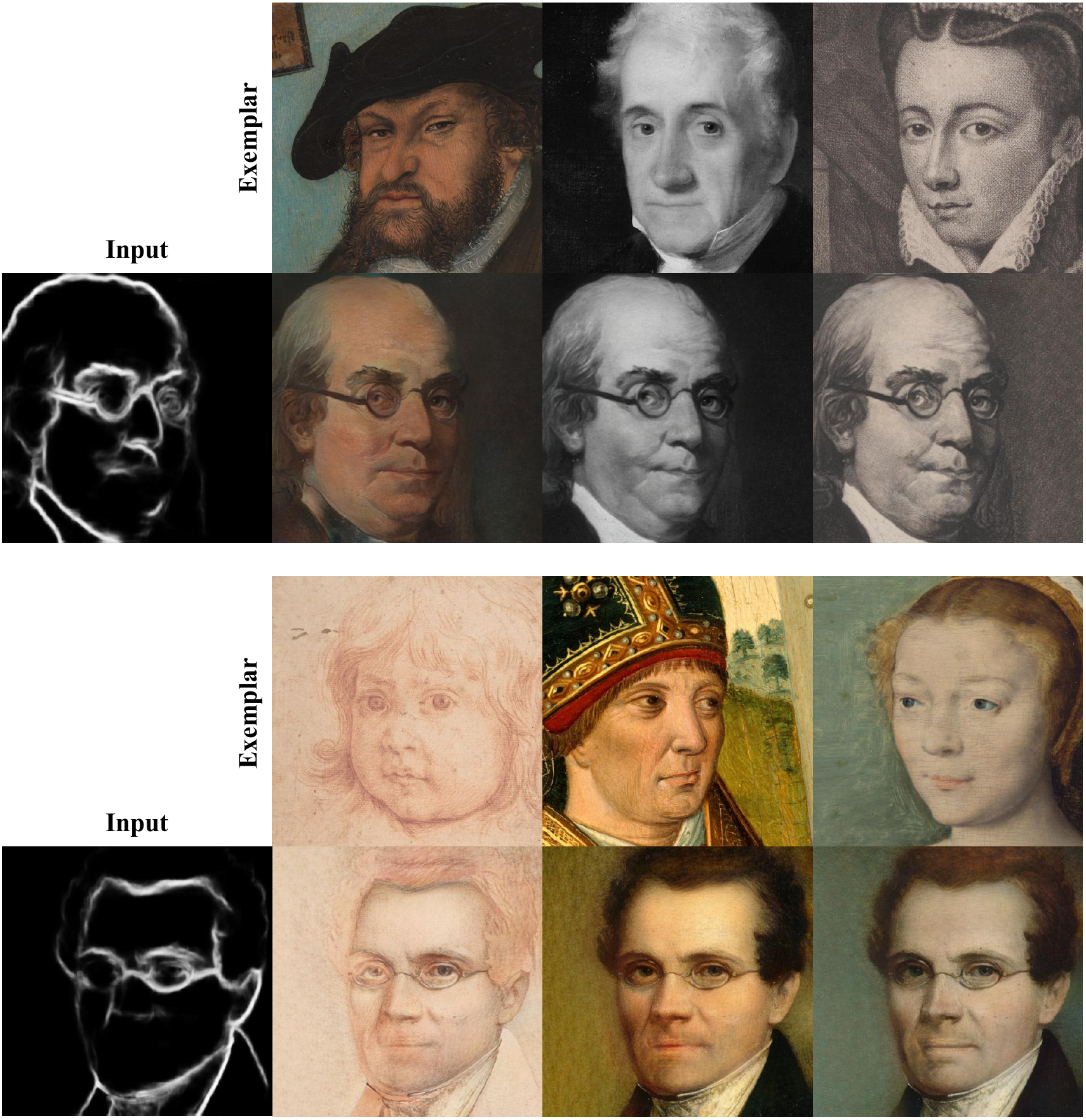}
\caption{Exemplar-based image translation results on MetFaces dataset at the resolution 512$\times$512.}
\label{fig:m1}
\end{figure}

\begin{figure}[t]
\centering
\includegraphics[width=0.9\textwidth]{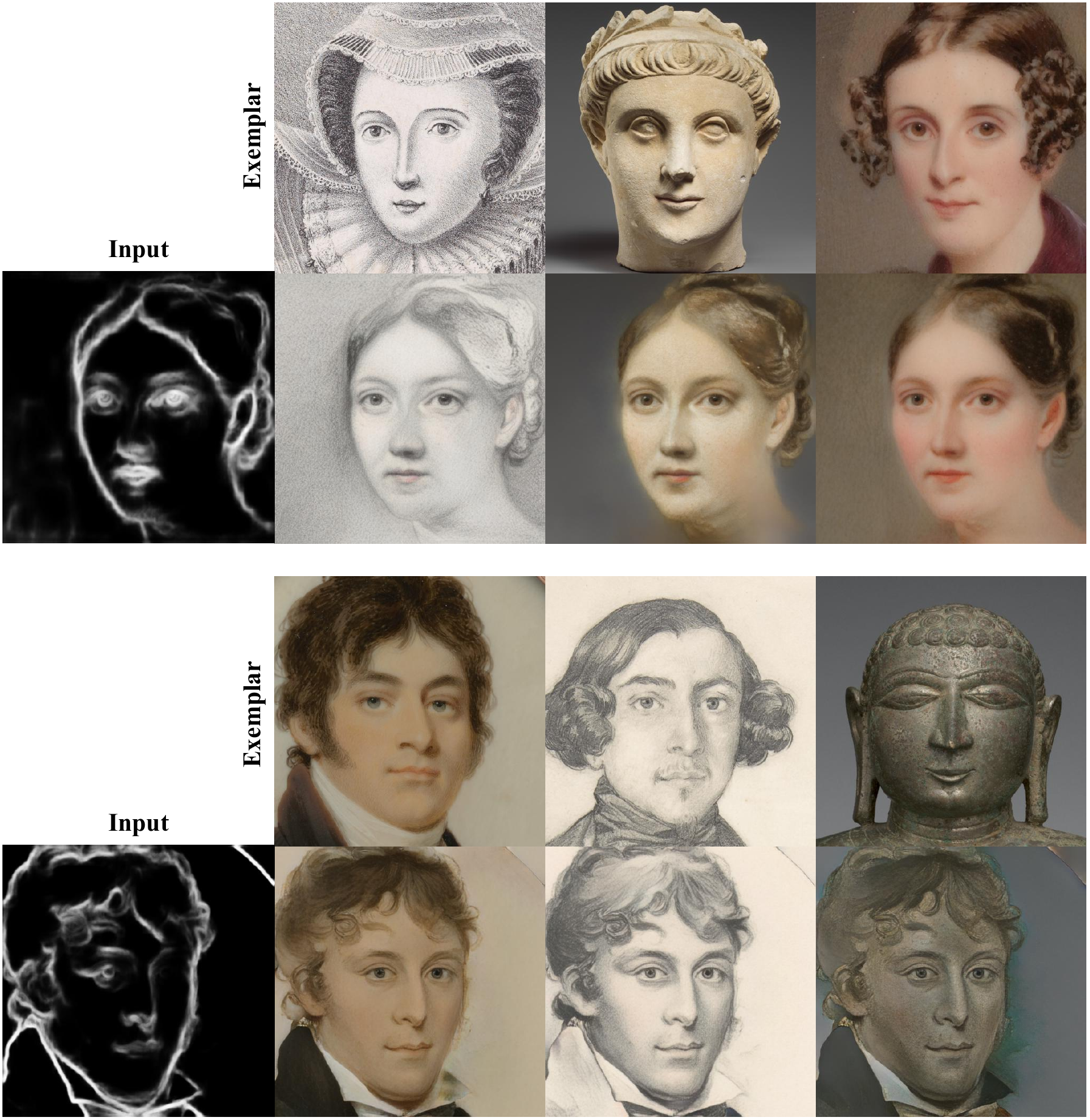}
\caption{Exemplar-based image translation results on MetFaces dataset at the resolution 512$\times$512.}
\label{fig:m2}
\end{figure}

Figure \ref{fig:ce1} and \ref{fig:ce2} show more translation results on the CelebA-HQ dataset. The results show that the skin tone of the face, the makeup, the hair color, and other styles are kept very fine, while the content of the generated image is strictly consistent with the input image. Similarly, Figure \ref{fig:m1} and \ref{fig:m2} also show more sketch-to-image translation results on the MetFaces dataset.

\clearpage

\begin{figure}[h]
\centering
\includegraphics[width=0.9\textwidth]{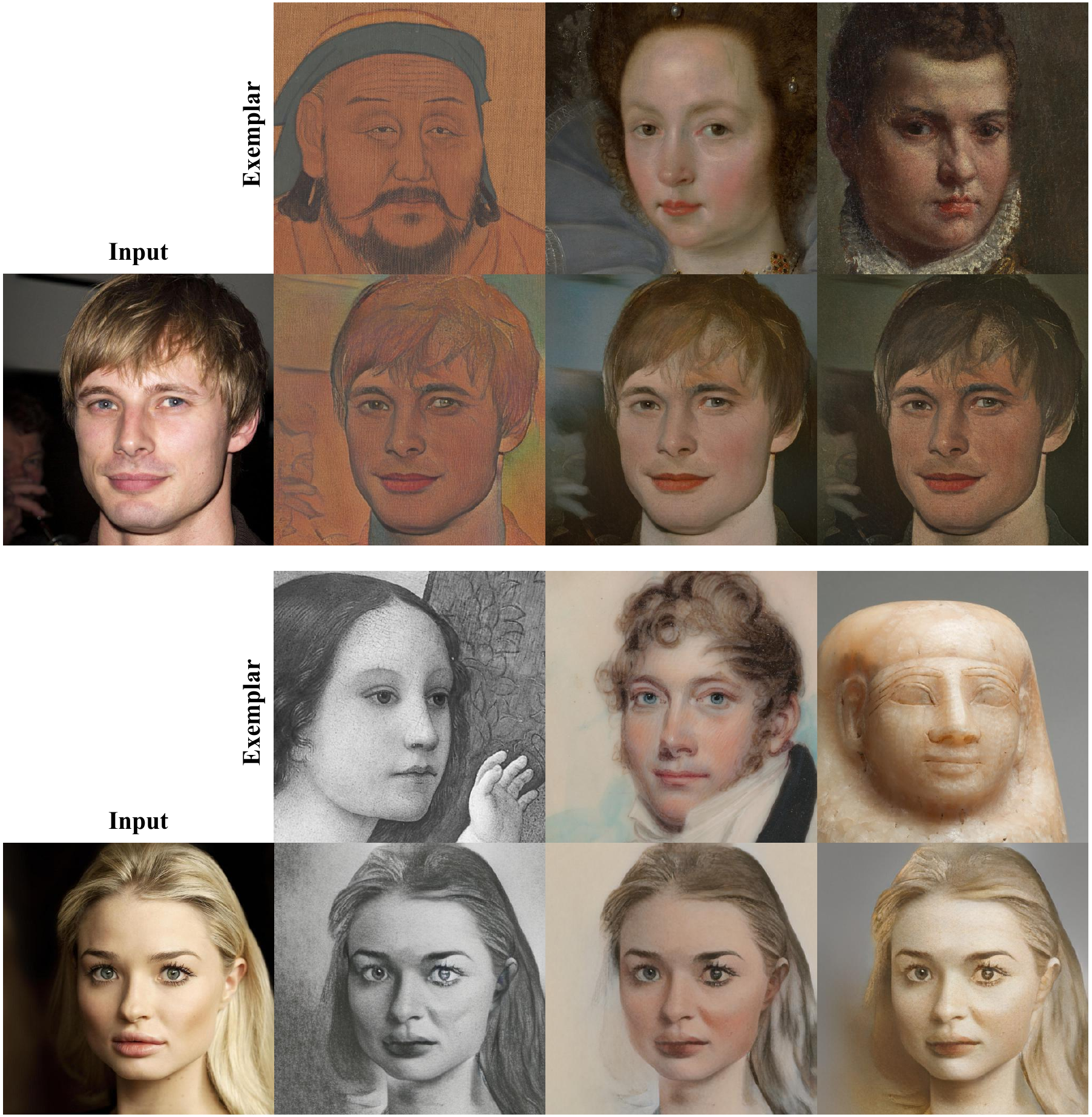}
\caption{Portrait stylization. CFFT-GAN can transfer image styles from the MetFaces dataset to the faces of the CelebA-HQ dataset. The model is only trained on the MetFaces dataset, and the model input is the edge image of the input face.}
\label{fig:c1}
\end{figure}

\begin{figure}[t]
\centering
\includegraphics[width=0.9\textwidth]{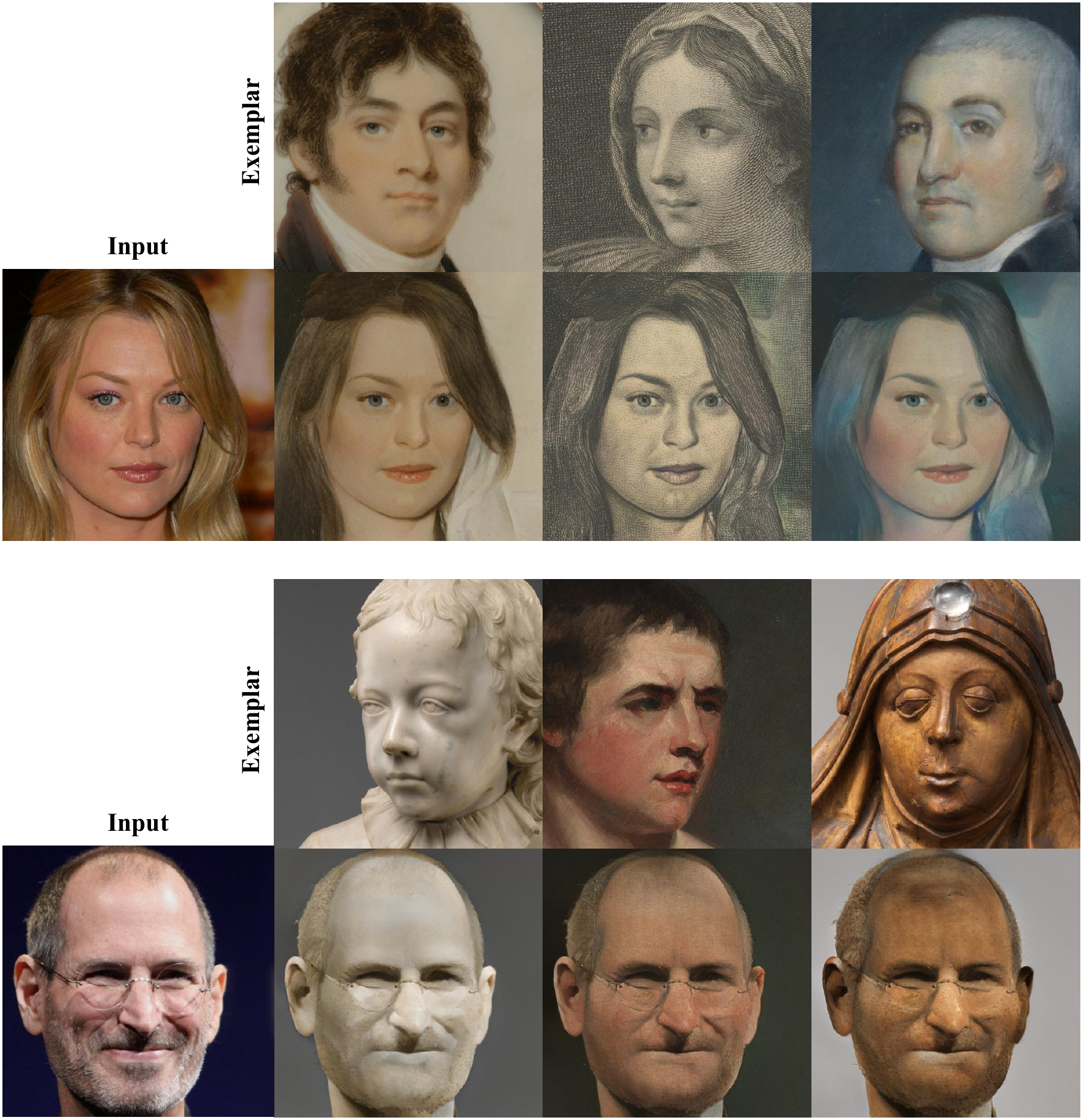}
\caption{Portrait stylization. CFFT-GAN can transfer image styles from the MetFaces dataset to the faces of the CelebA-HQ dataset. The model is only trained on the MetFaces dataset, and the model input is the edge image of the input face.}
\label{fig:c2}
\end{figure}

We show more cross-dataset image translation results, as shown in Figure\ref{fig:c1} and \ref{fig:c2}. The figures demonstrate the beautiful portrait stylization results. Our model is trained only on MetFaces and uses the face edges from the CelebA-HQ dataset as inputs, which demonstrate the good generalization of our method.

\clearpage

\begin{figure}[ht]
\centering
\includegraphics[width=0.65\textwidth]{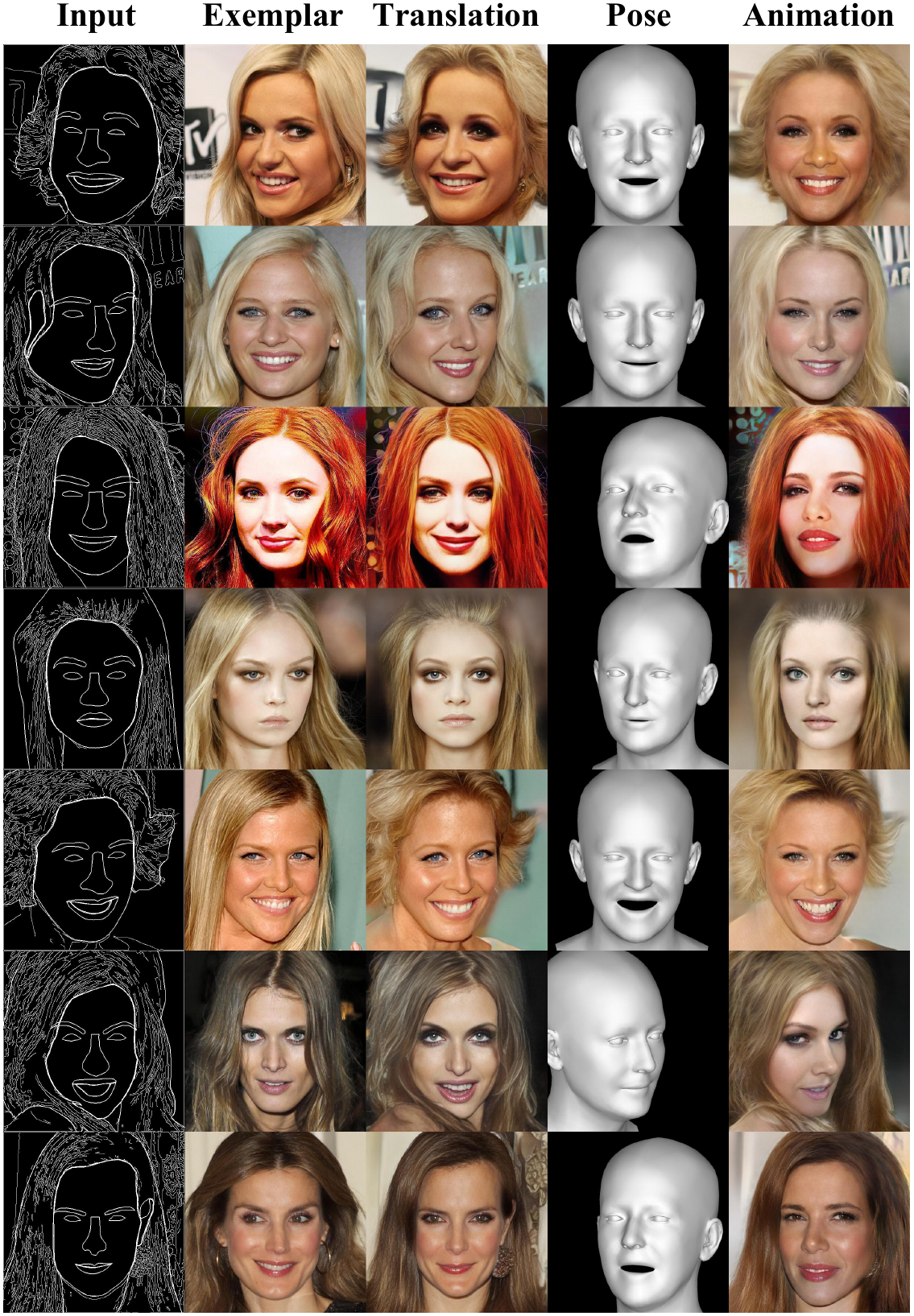}
\caption{Multi-domain image translation results. Columns 1, 2, and 4 are input, and columns 3 and 5 are output. The animation results are generated based on the translation results.}
\label{fig:ani1}
\end{figure}

Figure \ref{fig:ani1} shows more multi-domain image translation results on the CelebA-HQ dataset. Our method can implement both image translation and face animation at the same time by cascading CFFT modules.

% \begin{figure}[t]
% \centering
% \includegraphics[width=0.85\textwidth]{animation_2.pdf}
% \caption{Image translation and face animation results. Input edge and exemplar, CFFT-GAN can perform image translation, and then input pose information, the model can conduct face animation based on the previous image translation.}
% \label{fig:ani2}
% \end{figure}

\newpage

\begin{figure}[ht]
\centering
\includegraphics[width=0.9\textwidth]{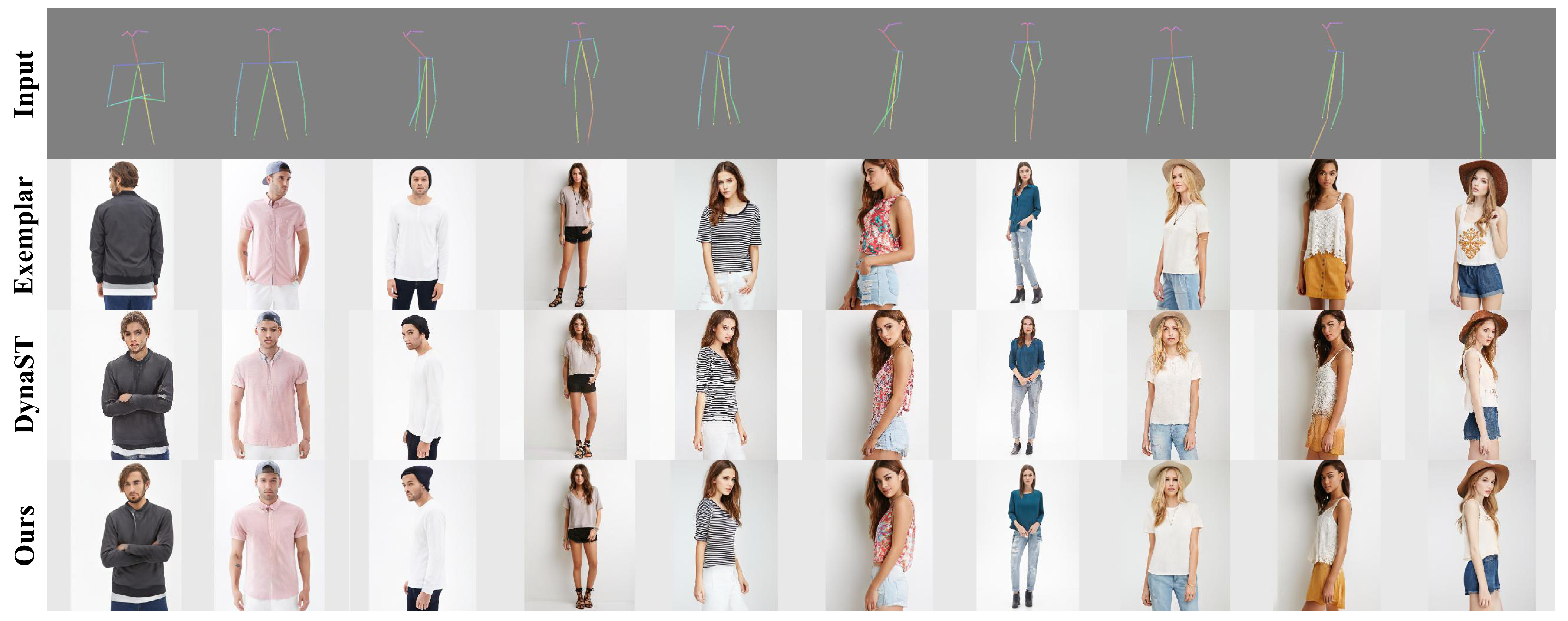}
\caption{Qualitative comparison results with transformer-based method Dynast.}
\label{fig:compare_DynaST}
\end{figure}

We compare a SOTA transformer-based method Dynast \cite{liu2022dynast} as shown in Figure \ref{fig:compare_DynaST}. Compared to that method, our approach is able to transfer more detailed features of the exemplar. 

\begin{figure}[ht]
\centering
\includegraphics[width=0.7\textwidth]{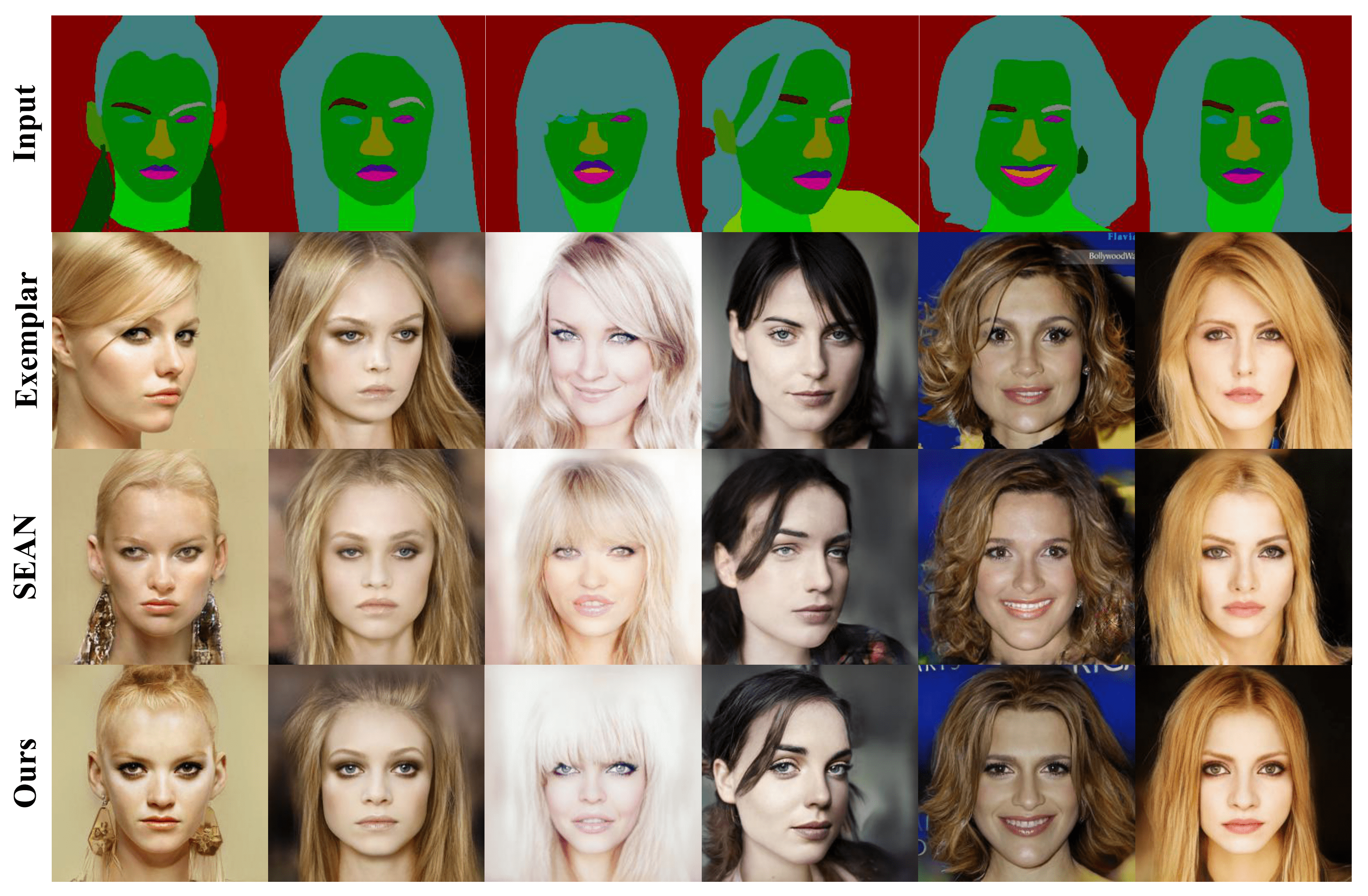}
\caption{Qualitative comparison results with SEAN.}
\label{fig:compare_SEAN}
\end{figure}

We also compare SEAN \cite{zhu2020sean} on CelebAHQ dataset as shown in Figure \ref{fig:compare_SEAN}, and our method is able to maintain the semantic structure of the input content image better. 

\newpage

\begin{figure}[ht]
\centering
\includegraphics[width=0.9\textwidth]{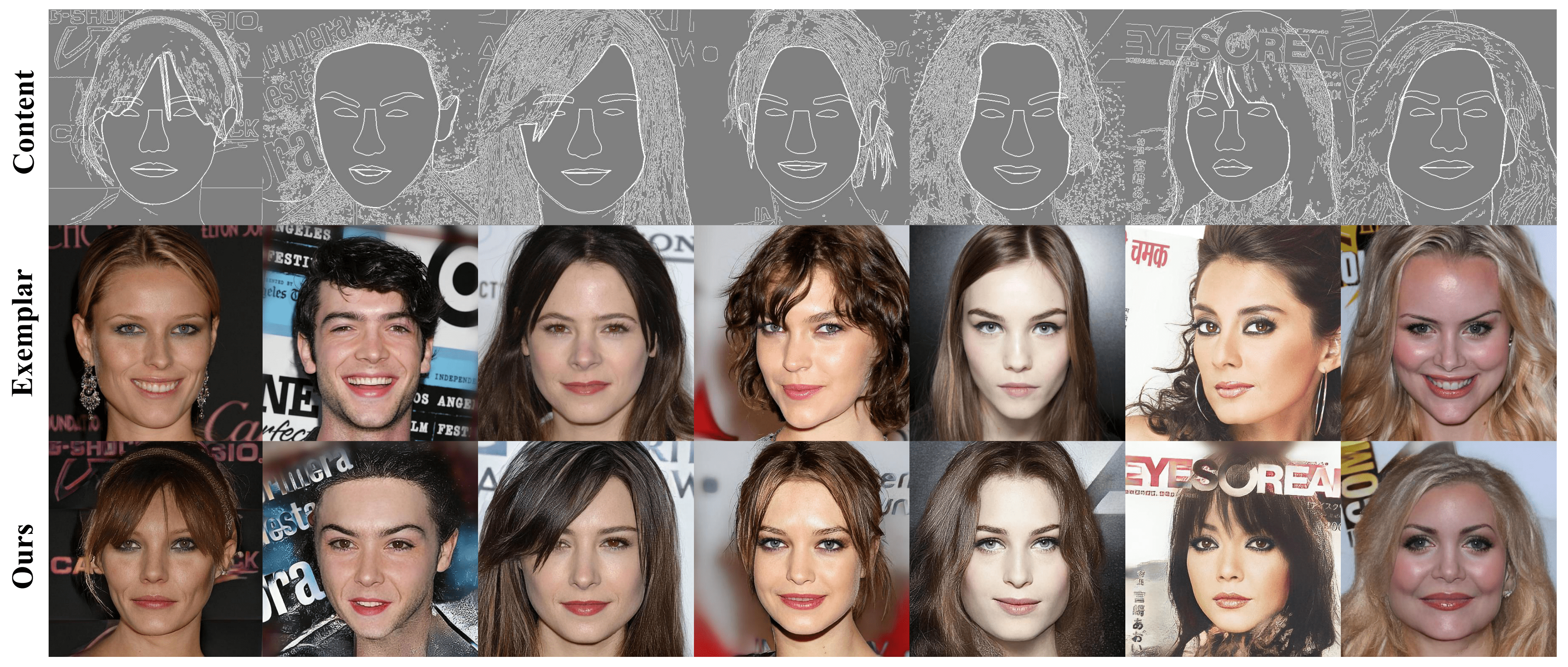}
\caption{Exemplar-based image translation results in complicated background.}
\label{fig:complicated_background}
\end{figure}

Our method is able to transfer the background information of the exemplar to the background structure of the content image. We add the background structure of the content images in the CelebAHQ dataset, and the translation results in Figure \ref{fig:complicated_background} show that our method is equally effective in transferring the background information from the exemplar to the content image.

\end{document}